\documentclass[]{techreport}

\newcommand{\model}{R2M-Bench}
\renewcommand{\reportshorttitle}{\model{}}
\renewcommand{\reportdate}{August 25, 2026}
\newcommand{\reportpdftitle}{R2M-Bench: Evaluating Revisit Memory via Relative Consistency in Interactive Video World Models}
\newcommand{\reportpdfauthor}{Qiwen Gu, Bingjie Gao, Rui Chen, Geng Li, Jifan Li, Qishuai Wen, Li Niu, Jing Tang, Xiangxiang Chu, Junqiao Zhao}

\setreportlogo{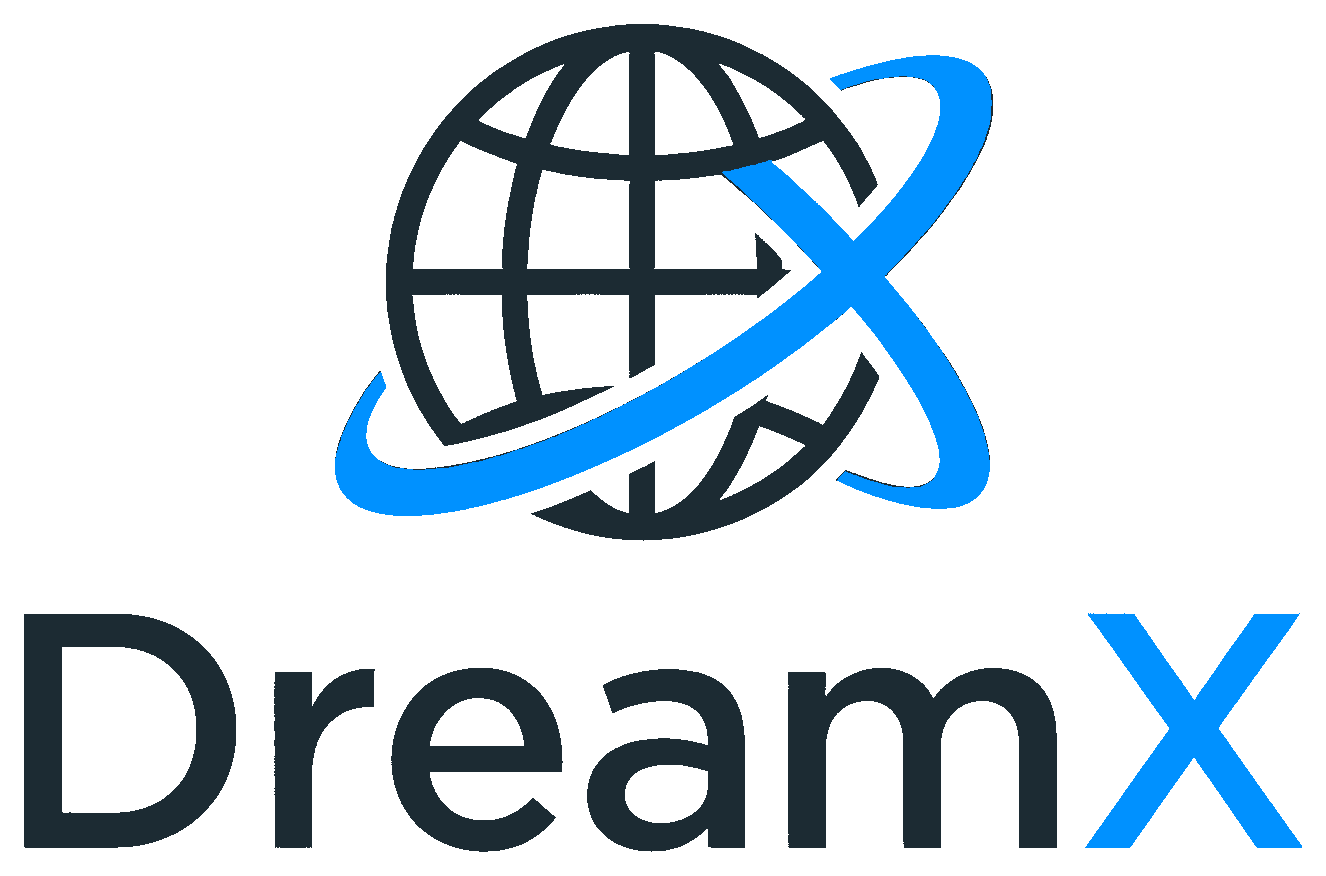}

\title{R2M-Bench: Evaluating Revisit Memory via Relative Consistency in Interactive Video World Models}
\author[1, 2, *, \dagger]{Qiwen Gu}
\author[1, 3, *, \dagger]{Bingjie Gao}
\author[1, \ddagger]{Rui Chen}
\author[1]{Geng Li}
\author[1]{Jifan Li}
\author[1]{Qishuai Wen}
\author[3]{Li~Niu}
\author[1, \S]{Jing Tang}
\author[1]{Xiangxiang Chu}
\author[2]{Junqiao Zhao}

\affiliation[1]{DreamX Team, Alibaba Group}
\affiliation[2]{Tongji University}
\affiliation[3]{Shanghai Jiao Tong University}

\contribution[*]{Equal contribution}
\contribution[\ddagger]{Project lead}
\contribution[\S]{Corresponding author}

\abstract{% !TEX root = ../iclr2026_conference.tex

High similarity between first-visit and return frames does not necessarily show that a video world model remembered the scene; the intervening rollout may simply have changed very little. This ambiguity makes absolute revisit scores sensitive to rendering stability, repetitive content, and failed motion. We introduce \emph{R2M-Bench} (\textbf{R}elative \textbf{R}evisit \textbf{M}emory Benchmark), a benchmark of observable revisit-selective consistency. For every detected return, R2M-Bench compares the revisit pair with two controls from the same rollout: a gap-matched non-revisit pair that measures generic temporal stability and a short-range pair that estimates short-horizon consistency. These comparisons produce \emph{MemoryGain} (MG), the revisit advantage over the temporal baseline, and the \emph{Normalized Memory Ratio} (NMR), which normalizes this advantage by the short-to-baseline dynamic range. R2M-Bench combines 100 reference scenes with three leave-and-return trajectories to form 300 instances and evaluates appearance fidelity, scene and object identity, local geometry, and persistent state. Across seven action-conditioned video world models, Overall NMR correlates with human consistency judgments at Spearman's $\rho=0.547$ (95\% CI $[0.45,0.63]$). Its within-model correlation magnitude with generated motion is $0.072$, compared with $0.207$ for raw revisit similarity, indicating that relative calibration substantially reduces the slow-motion shortcut. DreamX-World-Memo achieves the highest Overall NMR among the evaluated video models. Together, these results support same-rollout relative calibration as a practical way to distinguish revisit-specific consistency from generic temporal stability.
}
\metadata[GitHub]{\url{https://github.com/AMAP-ML/R2MBench}}
\metadata[Contact]{\email{guangyu.tj@alibaba-inc.com}}
\usepackage{amsmath}
\usepackage{amssymb}
\usepackage{array}
\usepackage{colortbl}
\usepackage{enumitem}
\usepackage{fancyhdr}
\usepackage{float}
\usepackage{xurl}

\definecolor{BrandPurple}{HTML}{7366CC}
\definecolor{BrandCyan}{HTML}{00B4E5}
\definecolor{LightGray}{HTML}{F5F5F5}
\definecolor{TableRowAlt}{HTML}{EAF4FB}
\definecolor{HeaderGray}{HTML}{888888}
\definecolor{TableRule}{HTML}{AEBBC5}

\newcommand{\reporttablehead}[1]{\textcolor{ReportText}{\textbf{#1}}}
\newcommand{\reporttablegroup}[1]{\textcolor{ReportPrimary}{\textbf{#1}}}
\newcommand{\reporttablepanelrow}{\rowcolor{ReportSurface}}
\newcommand{\reporttablegrouprow}{\rowcolor{ReportSurface!55}}
\newcommand{\reporttablehighlightrow}{\rowcolor{ReportSurface}}

\newcolumntype{L}[1]{>{\raggedright\arraybackslash}m{#1}}
\newcolumntype{C}[1]{>{\centering\arraybackslash}m{#1}}

\newtcolorbox{highlight}{
    colback=BrandCyan!5,
    colframe=BrandCyan,
    arc=2pt,
    boxrule=0.8pt,
    left=6pt, right=6pt, top=4pt, bottom=4pt,
    breakable
}

\newtcolorbox{keyfinding}{
    colback=BrandPurple!5,
    colframe=BrandPurple,
    coltitle=white,
    fonttitle=\bfseries,
    title=Key Finding,
    arc=2pt,
    boxrule=1pt,
    left=6pt, right=6pt, top=4pt, bottom=4pt,
    breakable
}

\newtcolorbox{tipbox}[1]{
    colback=ReportSurface,
    colframe=ReportPrimary,
    coltitle=white,
    fonttitle=\bfseries,
    title=#1,
    arc=2pt,
    boxrule=1pt,
    left=6pt, right=6pt, top=4pt, bottom=4pt,
    breakable
}

\fancypagestyle{plain}{%
    \fancyhf{}
    \fancyhead[L]{\footnotesize\color{HeaderGray}\sffamily \reportshorttitle}
    \fancyhead[R]{\footnotesize\color{HeaderGray} \thepage}
    \fancyfoot[C]{}

}

\fancypagestyle{firstpage}{%
    \fancyhf{}
    \fancyhead[L]{\small\sffamily\color{ReportPrimary} DreamX Team}
    \fancyhead[R]{\small\color{HeaderGray} \reportdate}
    \fancyfoot[C]{\footnotesize\color{gray}\thepage}

}

\usepackage{amsmath,amsfonts,bm}

\def\eqref#1{equation~\ref{#1}}
\def\1{\bm{1}}

\DeclareMathAlphabet{\mathsfit}{\encodingdefault}{\sfdefault}{m}{sl}
\SetMathAlphabet{\mathsfit}{bold}{\encodingdefault}{\sfdefault}{bx}{n}

\usepackage{multirow}
\usepackage{fvextra}

\newif\ifarxivversion
\arxivversiontrue

\DeclareUnicodeCharacter{2014}{--}
\DeclareUnicodeCharacter{2212}{-}
\DeclareUnicodeCharacter{2264}{<=}

\hypersetup{
  colorlinks=true,
  pdftitle={\reportpdftitle},
  pdfauthor={\reportpdfauthor},
  pdfsubject={Benchmarking revisit-selective consistency in interactive video world models},
  pdfkeywords={world models, video generation, revisit memory, evaluation benchmark}
}

\begin{document}

\maketitle
\thispagestyle{firstpage}

\begingroup
\renewcommand{\thefootnote}{\fnsymbol{footnote}}
\footnotetext[2]{Work done during an internship at DreamX Team, Alibaba Group.}
\endgroup

\begin{center}
\captionsetup{type=figure}
\includegraphics[width=0.96\textwidth]{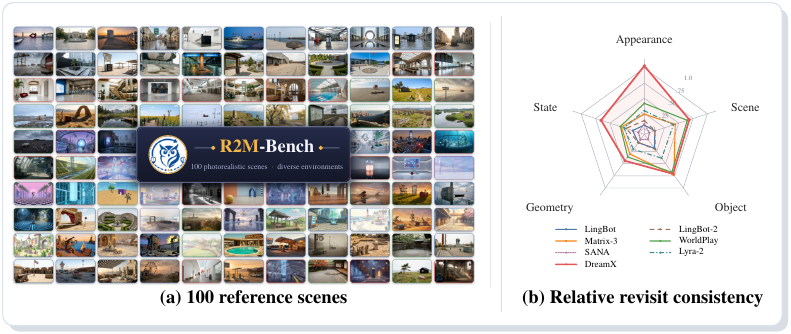}
\captionof{figure}{
\textbf{R2M-Bench at a glance.}
\textbf{(a)} The 100 reference scenes, each paired with three leave-and-return trajectories, define 300 benchmark instances.
\textbf{(b)} Template-balanced relative revisit consistency across five evaluation families for seven video world models; larger values indicate stronger recovery on return.
}
\label{fig:teaser}
\end{center}

% !TEX root = ../iclr2026_conference.tex

\section{Introduction}

World models provide predictive environments for planning, simulation, and embodied decision making. Video generation offers a practical route to such environments because it can synthesize high-dimensional observations directly, and recent action- or camera-conditioned models support increasingly long, controllable rollouts~\citep{yang2024cogvideox,hunyuanvideo2024,wan2025,zheng2024opensora,bruce2024genie,alonso2024diamond,wu2024ivideogpt}. Visual plausibility alone, however, is not enough for interaction. When an agent leaves a room, street corner, or object arrangement and later returns, the generated world should recover the same place rather than produce another plausible scene. A commanded revisit therefore tests whether the model preserves scene identity, object state, layout, and local structure through an intervening rollout. Figure~\ref{fig:teaser} previews R2M-Bench, from its reference-scene and trajectory construction to its template-balanced comparison of seven video world models.

Evaluation has not yet cleanly isolated this capability. General video benchmarks measure perceptual quality, temporal smoothness, prompt alignment, and physical plausibility~\citep{huang2023vbench,huang2025vbench2,liu2024evalcrafter}. World-model benchmarks additionally examine controllability, interactive response, and long-horizon generation~\citep{Li2025WorldModelBench,duan2025worldscore,wu2026omni}. Recent memory-oriented benchmarks consider revisit frames or long-horizon state retention~\citep{zhang2026mbench,ye2026mind}. These efforts make persistent-world failures visible, but an absolute similarity score between a first visit and a return still leaves an important ambiguity: how much of the measured similarity is specific to the revisit, and how much simply reflects the rollout's general visual stability?

\begin{figure}[!t]
\centering
\includegraphics[width=\textwidth]{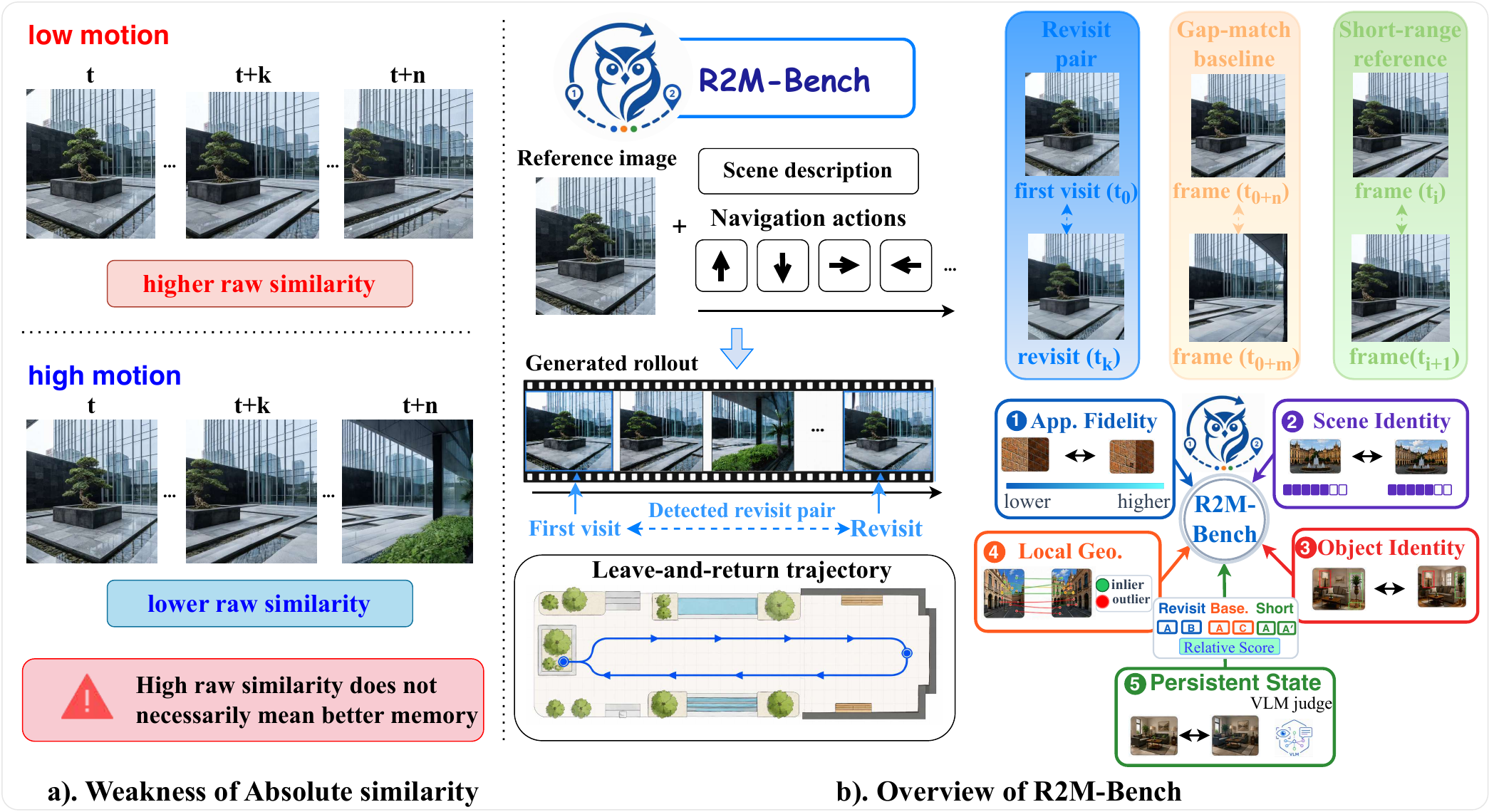}
\caption{
\textbf{Overview of R2M-Bench.}
(a) Absolute first-visit/revisit similarity is not sufficient evidence of memory because motion magnitude and rendering stability can inflate the raw score.
(b) R2M-Bench evaluates navigation-style generated rollouts by detecting leave-and-return revisit pairs from commanded trajectories, then compares each revisit pair with a gap-matched baseline and a short-range reference and reports MemoryGain and NMR across the metric families.
}
\label{fig:overview}
\end{figure}

Absolute revisit similarity is particularly difficult to compare across current world models. Even under the same commanded trajectory, models differ in rendering quality, motion magnitude, camera-execution speed, and the extent to which the generated view actually changes. As Figure~\ref{fig:overview}(a) illustrates, a low-motion rollout can achieve high first-visit/return similarity because most frames remain alike; in the limiting case, the camera never visibly leaves the initial view. Larger camera motion can lower the same score even when a model successfully returns. A revisit benchmark should therefore not reward sharp or conservative rendering that is equally present at ordinary times in the rollout.

We operationalize model revisit memory as \emph{observable revisit-selective consistency}. After a commanded return, the generated observation should recover the previously visited scene and state more consistently than comparable non-revisit observations from the same video. This definition concerns model behavior. It neither assumes that the model contains an explicit memory module nor attributes performance to a particular internal mechanism. The distinction is important because systems with different architectures, context policies, and retrieval strategies can exhibit similar output-level behavior.

R2M-Bench implements this definition through same-video temporal calibration (Figure~\ref{fig:overview}(b)). Starting from a reference image and a navigation script, the benchmark evaluates the generated rollout and mines frame pairs corresponding to commanded returns. Each revisit pair is compared with two references from the same video. A gap-matched non-revisit pair estimates the model's ordinary self-similarity over a comparable temporal interval. A short-range pair estimates the consistency available over nearby frames, before long-horizon drift dominates. \emph{MemoryGain} (MG) reports the direct advantage of the revisit pair over the gap-matched baseline. The \emph{Normalized Memory Ratio} (NMR) divides this advantage by the short-to-baseline dynamic range, yielding a dimensionless score that can be compared across metrics. Because all three pair types come from the same rollout, slow or nearly static generations raise the baseline as well as the revisit score and receive little selective advantage.

For broader context, Table~\ref{tab:benchmark_comparison} situates R2M-Bench among representative video-generation, world-model, and memory-oriented benchmarks by comparing the evaluation-design properties most relevant to revisit memory.

The benchmark combines 100 reference scenes with three leave-and-return trajectory templates, producing 300 benchmark instances. The trajectories include a direct out-and-back return, a translation followed by rotation, and a longer closed loop. Together they expose failures caused by viewpoint change, accumulated control error, autoregressive drift, and context attenuation. The reference set spans indoor, urban, natural, and abstract environments, with substantial variation in object density and scene structure.

R2M-Bench evaluates five complementary families of observable consistency. \textbf{Appearance Fidelity} captures changes in color, texture, and low-level perceptual appearance. \textbf{Scene Identity} tests whether the return remains recognizable as the same place under moderate viewpoint or rendering changes. \textbf{Object Identity} measures whether salient objects remain present with compatible appearance and semantics. \textbf{Local Geometry} tests whether local structures still support feature correspondence and a common two-view geometry. \textbf{Persistent State} assesses whether the pair depicts a compatible scene state after accounting for camera motion and occlusion. Applying the same relative protocol to all five families exposes partial failures that a single similarity metric would hide, such as preserving scene semantics while changing the objects or spatial layout.

We evaluate seven action-conditioned video world model variants and report a separate 3D rendering reference. On 210 sampled rollouts, Overall NMR correlates with human consistency judgments at Spearman's $\rho=0.547$, with a 95\% bootstrap interval of $[0.45,0.63]$. An independent optical-flow diagnostic finds a substantially smaller correlation magnitude with generated motion for NMR than for raw revisit similarity ($0.072$ versus $0.207$). DreamX-World-Memo achieves the highest Overall NMR among the evaluated video models, followed by HY-WorldPlay and Matrix-Game 3.0, while the trajectory-wise profiles show that even the strongest systems degrade on longer closed-loop returns. Retrieval-based systems occupy the top tier in this evaluation, but differences in architecture, scale, training, and inference preclude a causal conclusion about retrieval itself.

\textbf{We make three contributions:}
\begin{itemize}
\item We introduce \emph{R2M-Bench}, a 300-instance benchmark that combines 100 diverse scenes with three complementary return trajectories and evaluates five dimensions of visual and state consistency.
\item We develop a same-video relative protocol that distinguishes revisit-selective recovery from generic temporal stability. MG reports the direct revisit advantage, while NMR normalizes this advantage for cross-metric and family-level aggregation.
\item We provide a multi-model empirical study with human validation, objective motion analysis, controlled motion and quality diagnostics, and trajectory-wise evaluation. These analyses characterize both the alignment and the remaining limitations of relative revisit scoring.
\end{itemize}

% !TEX root = ../iclr2026_conference.tex

\section{Related Work}
\label{sec:related}

\textbf{Video generation and long-horizon rollouts.}
Video generation has moved from early adversarial and autoregressive models for spatio-temporal generation and discrete visual-token prediction~\citep{saito2017tgan,tulyakov2018mocogan,yan2021videogpt} to diffusion-based video synthesis that extends image diffusion to the temporal domain~\citep{ho2022video,singer2022makeavideo,blattmann2023stable}. Recent transformer-based video foundation models improve visual fidelity, prompt following, motion quality, and temporal smoothness~\citep{yang2024cogvideox,hunyuanvideo2024,wan2025}. Since many video generators are trained on fixed-length clips, long videos are commonly produced through autoregressive or streaming rollouts, where future clips are conditioned on previously generated content. Recent methods reduce rollout degradation with self-forcing objectives, rolling or streaming denoising, and causal distillation recipes~\citep{huang2025selfforcing,liu2025rolling,yang2025longlive}. These methods provide the generative backbone for interactive video world models, but longer video alone does not ensure that the generated environment stays persistent.

\textbf{Action-driven interactive video world models and memory mechanisms.}
Recent work has pushed world models toward action-driven interactive video generation, where future observations respond to user actions, camera controls, keyboard--mouse inputs, navigation commands, or event instructions. Earlier world models studied environment dynamics for planning and control~\citep{ha2018world,lecun2022path}. Recent generative simulators and action-conditioned video models scale this idea to robotics, driving, games, and interactive environments~\citep{hu2023gaia,bruce2024genie,oasis2024,sun2025worldplay,robbyant2026advancing,shen2026lyra2,wang2026matrixgame3,mao2026yume1,wu2026infinite,team2026dreamx}. Unlike standalone video generators, these models support continuous exploration: an agent can move, turn, revisit locations, and interact over time.

% !TEX root = ../iclr2026_conference.tex

\begin{table}[!t]
\centering
\begingroup
\caption{
Comparison with representative video-generation, world-model, and memory-oriented benchmarks from the perspective of evaluation design.
\textbf{Long}: long-horizon or multi-turn evaluation;
\textbf{Mem.}: explicit memory or persistent-state evaluation;
\textbf{Rev.}: constructed revisit or loop-closure test cases;
\textbf{View-Rob.}: robustness to moderate viewpoint deviations;
\textbf{Rel. Score}: relative scoring against within-video temporal controls rather than absolute similarity;
\textbf{Multi-axis}: evaluation across multiple consistency levels.
\checkmark denotes an explicit benchmark design goal;
$\triangle$ denotes partial or indirect coverage.
}
\label{tab:benchmark_comparison}
\scriptsize
\setlength{\tabcolsep}{5.5pt}
\arrayrulecolor{TableRule}
\resizebox{0.96\textwidth}{!}{
\begin{tabular}{@{}lcccccc@{}}
\toprule
\reporttablehead{Benchmark} & \reporttablehead{Long} & \reporttablehead{Mem.} & \reporttablehead{Rev.} & \reporttablehead{View-Rob.} & \reporttablehead{Rel. Score} & \reporttablehead{Multi-axis} \\
\midrule
VBench~\citep{huang2023vbench}
& -- & -- & -- & -- & -- & \checkmark \\

EvalCrafter~\citep{liu2024evalcrafter}
& -- & -- & -- & -- & -- & \checkmark \\

WorldModelBench~\citep{Li2025WorldModelBench}
& -- & -- & -- & -- & -- & \checkmark \\

WorldScore~\citep{duan2025worldscore}
& \checkmark & $\triangle$ & -- & -- & -- & \checkmark \\

WBench~\citep{ying2026wbench}
& \checkmark & $\triangle$ & -- & -- & -- & \checkmark \\

Omni-WorldBench~\citep{wu2026omni}
& $\triangle$ & -- & -- & -- & -- & \checkmark \\

MBench~\citep{zhang2026mbench}
& \checkmark & \checkmark & \checkmark & $\triangle$ & -- & \checkmark \\

MIND~\citep{ye2026mind}
& \checkmark & \checkmark & \checkmark & \checkmark & -- & \checkmark \\
\midrule
\reporttablehighlightrow
\textbf{R2M-Bench (Ours)}
& \textbf{\checkmark} & \textbf{\checkmark} & \textbf{\checkmark} & \textbf{\checkmark} & \textbf{\checkmark} & \textbf{\checkmark} \\
\bottomrule
\end{tabular}
}
\arrayrulecolor{black}
\endgroup
\end{table}

\begin{figure}[!t]
\centering
\includegraphics[width=0.99\linewidth]{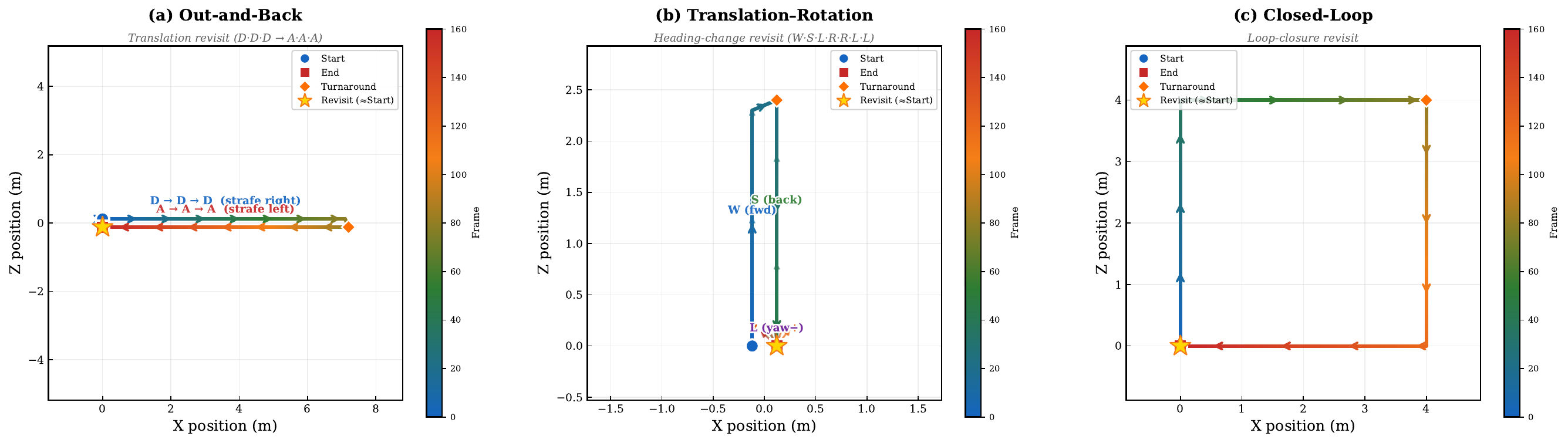}
\caption{%
\textbf{Bird's-eye view of the three trajectory templates used in R2M-Bench.}
Color indicates temporal progression from start (\textcolor{blue}{blue}) to end (\textcolor{red}{red}), and arrows indicate camera motion.
\textbf{(a)}~\emph{Out-and-back}: the agent moves away and reverses to the starting pose.
\textbf{(b)}~\emph{Translation--rotation}: the agent revisits the starting region after heading changes.
\textbf{(c)}~\emph{Closed-loop}: the agent traverses a loop and returns to the starting pose.
The $\star$ marker denotes the revisit point.
}
\label{fig:trajectory-templates}
\end{figure}

The three controlled return patterns used to instantiate the benchmark are summarized in Figure~\ref{fig:trajectory-templates} and detailed in Section~\ref{sec:data}.

Long interactive rollouts need memory. Existing mechanisms include \emph{context replay or retrieval}, which conditions generation on selected past frames, latents, or retrieved context~\citep{sun2025worldplay,hong2025relic,hu2026longliverag,peng2026compretrieval}; \emph{compressed memory}, which stores long histories as fixed-budget tokens, object slots, or summaries~\citep{kim2026memrope,dou2026slotmemory,wu2026infinite}; \emph{state memory}, which maintains a compact latent state across chunks~\citep{chen2025rad,yu2025videossm,li2026linearatt,zhu2026sanawm}; and \emph{spatial or geometric memory}, which organizes past views in 3D or view-indexed structures~\citep{li2025vmem,wang2026mirage,shen2026lyra2}. These mechanisms can reduce drift and improve temporal coherence, but they do not guarantee that a previously observed place remains persistent after leaving the field of view.

\textbf{Evaluation of video/world models and revisit memory.}
Video generation benchmarks have expanded from generic quality assessment to multi-dimensional evaluation of visual quality, temporal stability, motion, text alignment, compositionality, physics, and long-video coherence~\citep{huang2023vbench,huang2024vbenchplus,huang2025vbench2,liu2024evalcrafter,sun2024t2vcomp,bansal2024videophy,meng2024phygen}. Complementary evaluation work targets perception-aligned motion quality~\citep{ling2025vmbench}, fine-grained entity-level video quality assessment~\citep{chen2025finger}, and narrative expression in long-video generation~\citep{feng2026narrlv}. World-model benchmarks evaluate instruction following, physical adherence, camera or action controllability, 3D consistency, interactive response, policy use, and long-horizon world generation~\citep{Li2025WorldModelBench,duan2025worldscore,quevedo2025worldgym,wu2026omni}. Recent memory-oriented benchmarks evaluate entity or environment consistency, action control, closed-loop revisits, and long-horizon state retention~\citep{zhang2026mbench,ye2026mind}. R2M-Bench focuses on calibrating the measurement of revisit consistency. Absolute first-visit/revisit similarity can be inflated by image quality, stable style, repetitive textures, slow motion, or near-static generation, so R2M-Bench measures whether similarity is selectively elevated at revisits relative to temporal references from the same generated video. It evaluates this calibrated revisit advantage through Appearance Fidelity, Scene Identity Preservation, Object Identity, Local Geometric Correspondence, and Persistent State Reasoning.

% !TEX root = ../iclr2026_conference.tex

\section{R2M-Bench Data Composition}
\label{sec:data}
R2M-Bench evaluates revisit memory in navigation-style interactive video world models. It pairs 100 reference scenes with each of three trajectory templates, yielding 300 benchmark instances. Each instance contains an initial reference image, a text prompt, and a canonical navigation script that induces a leave-and-return interaction. After generation, we retain the resulting video together with its reference image, prompt, script, and commanded trajectory for evaluation. Figure~\ref{fig:teaser}(a) shows the complete reference set. Its mutually exclusive primary partition contains 24 indoor, 44 urban, 28 natural, and 4 abstract scenes, while the non-exclusive attributes show that 87 scenes are object-rich and 76 are outdoor, with overlapping urban, natural, and indoor tags.

\paragraph{\bf{Trajectory Construction.}}
Each trajectory is a model-agnostic navigation script composed of simple action primitives, including translations along the forward--backward and left--right axes and left/right yaw rotations. We convert the same script to the input format required by each model: discrete actions for action-only interfaces, or camera deltas/poses for camera-conditioned interfaces. For analysis, we integrate the primitives into a commanded pose sequence, which records the intended trajectory and is later used to mine revisit pairs.

Figure~\ref{fig:trajectory-templates} shows the three trajectory templates: \emph{out-and-back} for near same-view revisits, \emph{translation--rotation} for viewpoint-changing revisits, and \emph{closed-loop} for loop-closure revisits. These templates cover common navigation patterns and test whether the generated world preserves local appearance, place identity, and global layout consistency after revisitation.

\paragraph{\bf{Initial Reference Selection.}}
For each trajectory, we select an initial reference image that defines the starting observation of the generated world. We use lightweight inclusion criteria: images should depict navigable scenes with recognizable spatial layout, stable visual anchors such as furniture, windows, road structures, signs, or distinctive objects, and enough free space for plausible camera motion. We include indoor, outdoor, urban, natural, and object-rich environments, while excluding close-ups, blank backgrounds, heavy text overlays, severe blur, and ambiguous scale. The selected references cover varied scenes while preserving the scene and object cues needed for memory evaluation.

\paragraph{\bf{Prompt Annotation.}}
Given the selected reference image, we annotate a text prompt that describes the initial environment and its salient visual content. We use Gemini-3.1-Pro-Preview to draft the description, then manually normalize it into a concise generation prompt. The prompt keeps persistent cues such as scene category, spatial layout, major objects, materials, and distinctive landmarks. We remove hallucinated objects, fragile exact counts, transient lighting details, and style-heavy wording, so the text conditions the starting scene without over-constraining revisit performance.

\FloatBarrier

% !TEX root = ../iclr2026_conference.tex

\section{Evaluation Method}
\label{sec:method}

\subsection{Relative Revisit Evaluation}

We treat revisit-memory evaluation as a relative comparison. Given a generated video $\mathbf{V}=(f_0,f_1,\ldots,f_{T-1})$ and its commanded pose sequence $\mathbf{P}=(p_0,p_1,\ldots,p_{T-1})$, absolute first-visit/revisit similarity is not enough evidence of memory. We instead compare each revisit pair with temporal references sampled from the same generated video. This comparison calibrates the revisit against the rollout's own visual and temporal behavior, reducing the extent to which image quality, rendering stability, or low motion alone can produce a high score.

For each video, we construct three frame-pair sets: revisit pairs $\mathcal{R}$, gap-matched baseline pairs $\mathcal{B}$, and short-range reference pairs $\mathcal{S}$. Revisit pairs test consistency at a commanded return to a previously visited region. Baseline pairs estimate the rollout's ordinary long-gap self-similarity under a comparable temporal separation. Short-range pairs measure local temporal consistency when little long-term retention is required.

\subsection{Frame-Pair Construction}

Let $p_t=(\mathbf{t}_t,\theta_t)$ denote the commanded camera state at frame $t$, where $\mathbf{t}_t$ is the camera position and $\theta_t$ is the yaw angle. We denote the circular yaw distance by $d_\theta(\theta_i,\theta_j)$.

\paragraph{Revisit pairs.}
A revisit pair is a temporally separated pair of frames whose commanded poses return to the same spatial region with a similar viewing direction. We mine revisit candidates from the commanded pose sequence:
\begin{equation}
\mathcal{R}
=
\left\{
(i,j)\ \middle|\
i<j,\
\|\mathbf{t}_i-\mathbf{t}_j\|_2 \leq \tau_{\mathrm{pos}},\
d_\theta(\theta_i,\theta_j) \leq \tau_{\mathrm{rot}},\
j-i \geq \max(0.2T,10)
\right\},
\label{eq:revisit_pairs}
\end{equation}
where $\mathbf{t}_i$ and $\theta_i$ denote the commanded position and yaw angle at frame $i$, $d_\theta(\cdot,\cdot)$ is the circular yaw distance, and $T$ is the effective video length after trimming static leading and trailing frames. The thresholds $\tau_{\mathrm{pos}}$ and $\tau_{\mathrm{rot}}$ are set adaptively from the commanded trajectory, as described in the implementation details. When multiple earlier frames satisfy the revisit condition for the same later frame, we keep the match with the smallest combined pose difference.

\paragraph{Gap-matched baseline pairs.}
For each revisit pair $(i,j)\in\mathcal{R}$ with temporal gap $\Delta=j-i$, we first construct candidate baseline pairs from the same video with a similar temporal gap:
\begin{equation}
\mathcal{C}_{(i,j)}
=
\left\{
(u,v)\ \middle|\ %
u<v,\ %
|(v-u)-\Delta|\leq \delta,\ %
(u,v)\notin \mathcal{R},\ %
u,v\notin \mathcal{N}_\eta(\mathcal{R})
\right\},
\label{eq:baseline_pairs}
\end{equation}
where $\delta$ is the gap tolerance and $\mathcal{N}_\eta(\mathcal{R})$ excludes frames within an $\eta$-frame neighborhood of revisit endpoints. We then sample one baseline pair $b_{(i,j)}\in\mathcal{C}_{(i,j)}$ for each revisit pair and define
\begin{equation}
\mathcal{B}
=
\left\{
b_{(i,j)}
\ \middle|\
(i,j)\in\mathcal{R},\ \mathcal{C}_{(i,j)}\neq\emptyset
\right\}.
\end{equation}

The baseline set is a same-video, temporal-gap-matched calibration reference. By assigning a matched non-target pair to each revisit pair, $\mathcal{B}$ estimates the similarity already present at ordinary long-gap moments under a comparable time budget. It is not intended to approximate a counterfactual rollout in which the model has no memory, nor does it control every difference in realized viewpoint or action execution. Its purpose is to expose generic high self-similarity: if a rollout moves slowly, remains nearly static, renders conservatively, or does not visibly leave the initial scene, baseline pairs also remain similar. We deliberately do not impose a spatial-far constraint, so these shortcuts raise the baseline rather than being excluded from calibration.

\paragraph{Short-range reference pairs.}
We also sample short-range pairs
\begin{equation}
\mathcal{S}
=
\left\{
(a,b)\ \middle|\ %
\Delta_{\min}\leq b-a\leq \Delta_{\max}
\right\},
\label{eq:short_pairs}
\end{equation}
where $\Delta_{\min}$ and $\Delta_{\max}$ define a small temporal window. These pairs measure local temporal consistency and provide a reference for the consistency that the model maintains when the time gap is short.

\subsection{MemoryGain and Normalized Memory Ratio}

We first define the scoring procedure for an arbitrary pairwise consistency measure $q$. Since different measures may have different directions, we convert $q$ into an oriented score $s_q$ such that larger values always indicate stronger consistency:
\begin{equation}
s_q(f_i,f_j)=
\begin{cases}
q(f_i,f_j), & \text{if higher } q \text{ indicates stronger consistency},\\
-q(f_i,f_j), & \text{if lower } q \text{ indicates stronger consistency}.
\end{cases}
\label{eq:oriented_score}
\end{equation}
For any pair set $\mathcal{P}\in\{\mathcal{R},\mathcal{B},\mathcal{S}\}$, we denote its average consistency score as
\begin{equation}
\bar{s}^{\mathcal{P}}_q
=
\frac{1}{|\mathcal{P}|}
\sum_{(i,j)\in\mathcal{P}}
s_q(f_i,f_j),
\qquad
\bar{s}^{\mathrm{rev}}_q=\bar{s}^{\mathcal{R}}_q,\quad
\bar{s}^{\mathrm{base}}_q=\bar{s}^{\mathcal{B}}_q,\quad
\bar{s}^{\mathrm{short}}_q=\bar{s}^{\mathcal{S}}_q.
\label{eq:mean_pair_score}
\end{equation}

\paragraph{MemoryGain.}
Our direct metric-level measure is \emph{MemoryGain}, defined as the revisit advantage over the gap-matched baseline:
\begin{equation}
\mathrm{MG}_q
=
\bar{s}^{\mathrm{rev}}_q
-
\bar{s}^{\mathrm{base}}_q.
\label{eq:memory_gain}
\end{equation}
MemoryGain asks whether consistency is selectively elevated at the commanded revisit relative to the rollout's ordinary long-gap similarity. Because both terms come from the same rollout, globally shared factors such as style, sharpness, conservative rendering, and low motion tend to affect both scores and are partially canceled by the difference. Positive MemoryGain therefore indicates a calibrated revisit advantage beyond generic temporal self-similarity; it should not be interpreted as a causal estimate of an internal memory mechanism.

\paragraph{Short-range dynamic range.}
To characterize the discriminability of each metric on a given rollout, we define the short-range dynamic range:
\begin{equation}
\mathrm{DR}_q
=
\bar{s}^{\mathrm{short}}_q
-
\bar{s}^{\mathrm{base}}_q.
\label{eq:dynamic_range}
\end{equation}
Dynamic range is not a memory score. It measures how much the metric separates local temporal consistency from the baseline. A small $\mathrm{DR}_q$ indicates a low-discriminability regime, such as near-static generation, repetitive scenes, or metrics that miss the relevant changes.

\paragraph{Normalized Memory Ratio.}
To compare and aggregate revisit advantages across metrics with different numerical scales, we define \emph{Normalized Memory Ratio} (NMR) as a dynamic-range-normalized score:
\begin{equation}
\mathrm{NMR}_q
=
\frac{\mathrm{MG}_q}{\mathrm{DR}_q+\epsilon}
=
\frac{
\bar{s}^{\mathrm{rev}}_q-\bar{s}^{\mathrm{base}}_q
}{
\bar{s}^{\mathrm{short}}_q-\bar{s}^{\mathrm{base}}_q+\epsilon
},
\label{eq:nmr}
\end{equation}
where $\epsilon$ is a small constant for numerical stability. NMR locates the revisit advantage relative to the metric's short-to-baseline dynamic range: a value of zero means that revisit consistency does not exceed the long-gap baseline, while a value of one means that it reaches the corresponding short-range consistency level. This normalization yields a dimensionless score that can be aggregated across metrics. For Overall NMR, we first average metrics within each family and then average the five family scores equally, preventing families with more metrics from receiving greater weight. However, NMR depends on both the revisit gain in the numerator and the dynamic range in the denominator. When $\mathrm{DR}_q$ is small, NMR can become unstable or amplified by a weak normalization scale. We therefore use MG as the direct evidence within each metric and NMR for family-level and overall aggregation, interpreting NMR jointly with MG and DR. Appendix~\ref{app:score_interpretation} discusses these cases in detail.

\subsection{Dimensions of Revisit Consistency}

Revisit consistency is not a single observable property: a model may preserve surface appearance while changing the identity of a place, retain recognizable objects while distorting their spatial arrangement, or produce two plausible views that cannot belong to the same persistent world. We therefore evaluate five complementary dimensions of revisit consistency. For each dimension, we identify the observable failure it is intended to expose, construct a pairwise test, and define the corresponding metric readouts. These dimensions characterize different aspects of the generated revisit; they should not be interpreted as distinct internal types of memory. All readouts are converted to oriented scores using Eq.~\ref{eq:oriented_score} and evaluated with the same MemoryGain and NMR protocol. Appendix~\ref{app:metric_details} gives the backbone configurations, preprocessing, and per-metric formulas.

\paragraph{\bf{Appearance Fidelity.}}
This dimension measures whether a revisit preserves the low-level visual appearance of the earlier observation. It exposes surface-level failures such as repainting, color or texture drift, blur, and changes in fine image structure. We directly compare the two frames using PSNR, SSIM, and LPIPS~\citep{zhang2018lpips}, which provide complementary pixel-, structure-, and perceptual-level readouts. A high revisit-relative score indicates that the model retains the earlier rendering beyond its generic temporal similarity; because these measures are sensitive to viewpoint and illumination, it does not by itself establish that the same place or world state has been recovered.

\paragraph{\bf{Scene Identity Preservation.}}
This dimension asks whether the returned observation remains recognizable as the same place under moderate appearance and viewpoint changes. A model fails when it generates a plausible scene that is semantically similar yet no longer corresponds to the previously visited location. We encode each frame independently with a general visual representation, DINOv2~\citep{oquab2023dinov2}, and with two place-recognition descriptors, BoQ~\citep{ali2024boq} and MutualVPR~\citep{gu2025mutualvpr}, then compute cosine similarity between the resulting global descriptors. The three similarities measure semantic scene resemblance and place recognizability from complementary representations. High revisit-relative similarity supports preserved location identity, but does not require exact object-level or geometric agreement.

\paragraph{\bf{Object Identity.}}
This dimension tests whether salient objects survive a revisit with consistent appearance and semantics. Typical failures include object disappearance, replacement, or identity drift even when the surrounding scene remains plausible. We detect and segment object anchors in the earlier frame using GroundingDINO~\citep{liu2023groundingdino} and SAM2~\citep{ravi2024sam2}, then re-detect each anchor concept independently in the paired frame rather than tracking it through the intervening video. Accepted candidates are compared through masked DINOv2 appearance features and CLIP~\citep{radford2021clip}-based semantic compatibility. We report appearance persistence and semantic persistence, assigning zero to unmatched anchors so that both identity quality and object survival affect the scores. High revisit-relative persistence indicates that the same object evidence is recovered, but does not ensure that the objects retain their original spatial arrangement.

\paragraph{\bf{Local Geometric Correspondence.}}
This dimension measures whether local structures in the two observations remain geometrically compatible. It targets failures in which a revisit preserves the scene category or object inventory but warps, rearranges, or replaces the underlying spatial structure. We extract SuperPoint keypoints~\citep{detone2018superpoint}, match them with LightGlue~\citep{lindenberger2023lightglue}, and test the matches with robust two-view geometry. The feature-correspondence ratio measures how much local evidence can be matched, while the RANSAC inlier ratio measures how much of that evidence supports a common epipolar geometry. High revisit-relative values provide local geometric support for a persistent scene, although they do not establish complete global 3D consistency.

\paragraph{\bf{Persistent State Reasoning.}}
This dimension asks whether paired observations depict the same local scene with state that remains consistent under camera motion and occlusion. It captures failures missed by lower-level measures, including scene replacement, contradictory layouts, and unexplained state changes. A vision-language model applies a fixed, pair-type-blind rubric to every revisit, baseline, and short-range pair and returns a scalar score. Higher revisit-relative scores indicate stronger persistent-state consistency than same-video controls, but not direct pixel or geometry alignment.

% !TEX root = ../iclr2026_conference.tex

\section{Experiments}
\label{sec:experiments}

\subsection{Implementation Details}

All evaluated video world models are run with chunked autoregressive inference: each rollout is generated one chunk at a time, and previous frames or model-specific memory states are fed back as context for the next chunk. Inference is performed on an 8-GPU NVIDIA H20 server. Evaluation videos contain approximately 481 frames, with small variations due to model-specific padding; we remove static leading and trailing frames and denote the remaining effective length by $T$.

For revisit-pair mining, we follow Eq.~\ref{eq:revisit_pairs} with $j-i\geq\max(0.2T,10)$ and adaptive thresholds $\tau_{\mathrm{pos}}=\mathrm{clip}(2P_{95}(\Delta t),0.1,1.0)$ meters and $\tau_{\mathrm{rot}}=\mathrm{clip}(2P_{95}(\Delta\theta),1.0,5.0)$ degrees. For each later frame, we keep the best earlier pose match. Baselines follow Eq.~\ref{eq:baseline_pairs}, using one gap-matched baseline per revisit pair with gap tolerance $\delta=\max(10,\mathrm{round}(0.3\Delta))$ and exclusion radius $\eta=\max(10,\lfloor T/20\rfloor)$. Short-range references follow Eq.~\ref{eq:short_pairs} with $\Delta_{\min}=2$, $\Delta_{\max}=\max(6,\lfloor0.02T\rfloor)$, and $N_{\mathrm{short}}=\min(100,\max(20,N_{\mathrm{rev}}))$. For NMR, we compute per-video MemoryGain and dynamic range, discard video and metric cases with $\mathrm{DR}\leq10^{-8}$, and take the ratio after averaging valid numerators and denominators.

For metric extraction, we use DINOv2-ViT-B/14 for generic scene features. SuperPoint uses at most 1024 keypoints with detection threshold $5\times10^{-4}$, and LightGlue uses filtering threshold 0.1. For vision-language judgment, we use Gemini-3.1-Pro-Preview with a fixed rubric; the full prompt appears in Appendix~\ref{app:vlm_prompt}.

\subsection{Benchmark Results}
\label{sec:exp_main}

\paragraph{\bf{Evaluation protocol.}}
We evaluate seven action-conditioned video world model variants: LingBot-World-Fast and LingBot-World-2-Fast~\citep{robbyant2026advancing,gao2026infiniteworldsversatileinteractions}, Matrix-Game 3.0~\citep{wang2026matrixgame3}, HY-WorldPlay~\citep{sun2025worldplay}, DreamX-World-Memo~\citep{team2026dreamx}, SANA-WM~\citep{zhu2026sanawm}, and Lyra-2~\citep{shen2026lyra2}. Each model is evaluated on all three trajectory templates, with up to 100 generated videos per model and trajectory and three runs with distinct random seeds. We additionally report WonderWorld~\citep{yu2024wonderworld}, an explicit 3D reconstruction and rendering system, as a near-perfect-memory reference rather than a directly comparable video model.

Table~\ref{tab:main_nmr_summary} separates two questions. The top panel reports metric-level MemoryGain (MG), which tests whether revisits are more consistent than gap-matched controls in the metric's native units. The bottom panel reports NMR, which normalizes that advantage by the same video's short-to-baseline dynamic range. Overall NMR averages metrics within each family and then weights the five families equally. We therefore use MG as direct evidence of a revisit advantage and NMR for comparisons across metrics and families.

\begin{table*}[!t]
\centering
\begingroup
\newcommand{\bestcell}[1]{\textbf{#1}}
\newcommand{\secondcell}[1]{\underline{#1}}
\newcommand{\refcell}[1]{\textcolor{HeaderGray}{#1}}
\newcommand{\refrow}[1]{\textcolor{HeaderGray}{#1}}
\caption{
Template-balanced MemoryGain (MG; top) and Normalized Memory Ratio (NMR; bottom), each averaged equally over Out-and-back, Translation--rotation, and Closed-loop.
Memory types: RC = raw context, FoV-R = field-of-view retrieval, G-R = geometric retrieval, and SSM = state-space recurrence.
WonderWorld (gray) is a separate 3D rendering reference.
For video models, \textbf{bold} and \underline{underlined} values mark the best and second-best results.
Overall NMR first averages metrics within each family and then the five families; MG is not aggregated because its units are metric-specific.
}
\label{tab:main_gain_summary}
\label{tab:main_nmr_summary}
\scriptsize
\setlength{\tabcolsep}{3.2pt}
\renewcommand{\arraystretch}{1.10}
\arrayrulecolor{TableRule}
\resizebox{\textwidth}{!}{
\begin{tabular}{@{}lllccccccccccc@{}}
\toprule
\reporttablepanelrow
\multicolumn{14}{c}{\reporttablegroup{MemoryGain (MG; higher is better)}} \\
\midrule
&
&
& \multicolumn{3}{c}{\reporttablegroup{Appearance Fidelity}}
& \multicolumn{3}{c}{\reporttablegroup{Scene Identity}}
& \multicolumn{2}{c}{\reporttablegroup{Object Identity}}
& \multicolumn{2}{c}{\reporttablegroup{Local Geometry}}
& \reporttablegroup{Persistent State} \\
\cmidrule(lr){4-6}
\cmidrule(lr){7-9}
\cmidrule(lr){10-11}
\cmidrule(lr){12-13}
\cmidrule(lr){14-14}
\reporttablehead{Method}
& \reporttablehead{Params}
& \reporttablehead{Mem. Type}
& \reporttablehead{PSNR}
& \reporttablehead{SSIM}
& \reporttablehead{LPIPS}
& \reporttablehead{DINO-sim}
& \reporttablehead{BoQ-sim}
& \reporttablehead{MVPR-sim}
& \reporttablehead{App. Persist}
& \reporttablehead{Sem. Persist}
& \reporttablehead{Feat. Corr.}
& \reporttablehead{Geo. Inlier}
& \reporttablehead{State Cons.} \\
\midrule
{\refrow{WonderWorld}}
& {\refrow{--}}
& {\refrow{Explicit 3D}}
& \refcell{12.828}
& \refcell{0.398}
& \refcell{0.548}
& \refcell{0.369}
& \refcell{0.633}
& \refcell{0.327}
& \refcell{0.254}
& \refcell{0.067}
& \refcell{0.535}
& \refcell{0.147}
& \refcell{2.676} \\
\midrule

LingBot-World-Fast
& 14B
& RC
& 0.283
& 0.018
& 0.007
& 0.029
& 0.041
& 0.027
& 0.022
& 0.012
& 0.040
& 0.008
& 0.186 \\

LingBot-World-2-Fast
& 14B
& RC
& 0.271
& 0.009
& 0.011
& 0.083
& 0.013
& 0.057
& 0.031
& 0.014
& 0.015
& 0.023
& 0.262 \\

Matrix-Game 3.0
& 5B
& FoV-R
& 1.039
& 0.024
& 0.060
& 0.145
& 0.176
& \bestcell{0.104}
& \bestcell{0.098}
& \bestcell{0.045}
& 0.145
& \secondcell{0.037}
& 0.476 \\

HY-WorldPlay
& 8B
& FoV-R
& \secondcell{2.357}
& \secondcell{0.055}
& \secondcell{0.155}
& \secondcell{0.146}
& \secondcell{0.208}
& 0.069
& 0.087
& \secondcell{0.029}
& \secondcell{0.180}
& 0.018
& \secondcell{0.510} \\

SANA-WM
& 2.6B
& SSM
& 0.393
& 0.008
& 0.009
& -0.004
& 0.003
& -0.019
& -0.001
& -0.003
& -0.001
& 0.009
& -0.008 \\

Lyra-2
& 14B
& G-R
& 1.318
& 0.041
& 0.090
& 0.129
& 0.134
& 0.086
& 0.058
& 0.022
& 0.096
& 0.031
& 0.422 \\

\reporttablehighlightrow
DreamX-World-Memo
& 5B
& FoV-R
& \bestcell{4.647}
& \bestcell{0.120}
& \bestcell{0.247}
& \bestcell{0.185}
& \bestcell{0.262}
& \secondcell{0.103}
& \secondcell{0.095}
& 0.027
& \bestcell{0.184}
& \bestcell{0.046}
& \bestcell{0.912} \\
\bottomrule
\end{tabular}
}
\par\vspace{6.0pt}
\setlength{\tabcolsep}{3.0pt}
\resizebox{\textwidth}{!}{
\begin{tabular}{@{}lllcccccccccccc@{}}
\toprule
\reporttablepanelrow
\multicolumn{15}{c}{\reporttablegroup{Normalized Memory Ratio (NMR; higher is better)}} \\
\midrule
&
&
& \multicolumn{3}{c}{\reporttablegroup{Appearance Fidelity}}
& \multicolumn{3}{c}{\reporttablegroup{Scene Identity}}
& \multicolumn{2}{c}{\reporttablegroup{Object Identity}}
& \multicolumn{2}{c}{\reporttablegroup{Local Geometry}}
& \reporttablegroup{Persistent State}
& \\
\cmidrule(lr){4-6}
\cmidrule(lr){7-9}
\cmidrule(lr){10-11}
\cmidrule(lr){12-13}
\cmidrule(lr){14-14}
\reporttablehead{Method}
& \reporttablehead{Params}
& \reporttablehead{Mem. Type}
& \reporttablehead{PSNR}
& \reporttablehead{SSIM}
& \reporttablehead{LPIPS}
& \reporttablehead{DINO-sim}
& \reporttablehead{BoQ-sim}
& \reporttablehead{MVPR-sim}
& \reporttablehead{App. Persist}
& \reporttablehead{Sem. Persist}
& \reporttablehead{Feat. Corr.}
& \reporttablehead{Geo. Inlier}
& \reporttablehead{State Cons.}
& \reporttablehead{Overall} \\
\midrule
{\refrow{WonderWorld}}
& {\refrow{--}}
& {\refrow{Explicit 3D}}
& \refcell{3.253}
& \refcell{3.076}
& \refcell{1.868}
& \refcell{1.110}
& \refcell{1.249}
& \refcell{1.096}
& \refcell{1.221}
& \refcell{1.319}
& \refcell{1.166}
& \refcell{1.132}
& \refcell{1.247}
& \refcell{1.510} \\
\midrule

LingBot-World-Fast
& 14B
& RC
& 0.047
& 0.083
& 0.030
& 0.073
& 0.080
& 0.103
& 0.126
& 0.245
& 0.098
& 0.043
& 0.103
& 0.100 \\

LingBot-World-2-Fast
& 14B
& RC
& 0.198
& 0.144
& 0.045
& 0.097
& 0.040
& 0.115
& 0.112
& 0.311
& 0.040
& 0.092
& 0.216
& 0.141 \\

Matrix-Game 3.0
& 5B
& FoV-R
& 0.257
& 0.192
& 0.251
& 0.449
& 0.400
& 0.470
& 0.572
& \bestcell{0.896}
& 0.385
& 0.226
& 0.300
& 0.403 \\

HY-WorldPlay
& 8B
& FoV-R
& \secondcell{0.446}
& \secondcell{0.308}
& \secondcell{0.488}
& \bestcell{0.734}
& \secondcell{0.589}
& \secondcell{0.626}
& \secondcell{0.602}
& 0.752
& \secondcell{0.488}
& \secondcell{0.243}
& \secondcell{0.318}
& \secondcell{0.485} \\

SANA-WM
& 2.6B
& SSM
& 0.094
& 0.071
& 0.051
& 0.002
& 0.012
& -0.049
& 0.016
& -0.007
& 0.002
& 0.020
& 0.006
& 0.016 \\

Lyra-2
& 14B
& G-R
& 0.284
& 0.286
& 0.306
& 0.457
& 0.330
& 0.435
& 0.337
& 0.442
& 0.236
& 0.180
& 0.255
& 0.310 \\

\reporttablehighlightrow
DreamX-World-Memo
& 5B
& FoV-R
& \bestcell{1.034}
& \bestcell{1.093}
& \bestcell{0.981}
& \secondcell{0.700}
& \bestcell{0.692}
& \bestcell{0.662}
& \bestcell{0.641}
& \secondcell{0.782}
& \bestcell{0.533}
& \bestcell{0.355}
& \bestcell{0.654}
& \bestcell{0.706} \\
\bottomrule
\end{tabular}
}
\arrayrulecolor{black}
\endgroup
\end{table*}

\paragraph{\bf{Overall comparison.}}
DreamX-World-Memo achieves the highest Overall NMR among the video models (0.706), followed by HY-WorldPlay (0.485), Matrix-Game 3.0 (0.403), and Lyra-2 (0.310). Its lead is supported by both components of the evaluation: DreamX-World-Memo has the largest MG on 8 of the 11 displayed metrics and the largest NMR on 9 of 11. The separate WonderWorld reference reaches 1.510 Overall NMR, consistent with its explicit 3D representation and deterministic rendering. NMR is not bounded by one; values above one indicate that the revisit advantage exceeds the corresponding short-to-baseline range and should be interpreted jointly with MG.

\paragraph{\bf{Dimension-specific profiles.}}
The aggregate ranking conceals distinct strengths. DreamX-World-Memo leads all three Appearance Fidelity metrics, two of three Scene Identity metrics, both Local Geometry metrics, and Persistent State. HY-WorldPlay is strongest on DINO scene similarity and ranks second overall, while Matrix-Game 3.0 has the highest semantic object-persistence score. LingBot-World-2-Fast improves its Overall NMR from 0.100 to 0.141 relative to LingBot-World-Fast, but the gains are uneven across families; SANA-WM remains near zero overall. Thus, no single appearance, semantic, or geometric metric adequately represents revisit consistency.

The four retrieval-based systems occupy the top four Overall NMR positions. Because model scale, training data, generation procedure, and inference also differ, this ranking is associative and does not isolate the effect of retrieval.

\begin{figure*}[!t]
\centering
\begin{minipage}{0.30\textwidth}
\centering
\includegraphics[width=\linewidth]{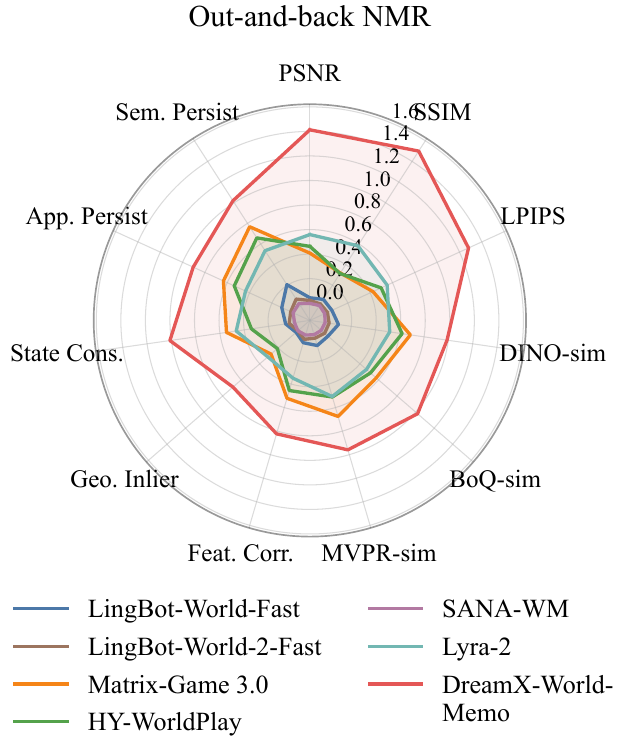}\\[-0.5em]
{\small (a) Out-and-back}
\end{minipage}
\hfill
\begin{minipage}{0.30\textwidth}
\centering
\includegraphics[width=\linewidth]{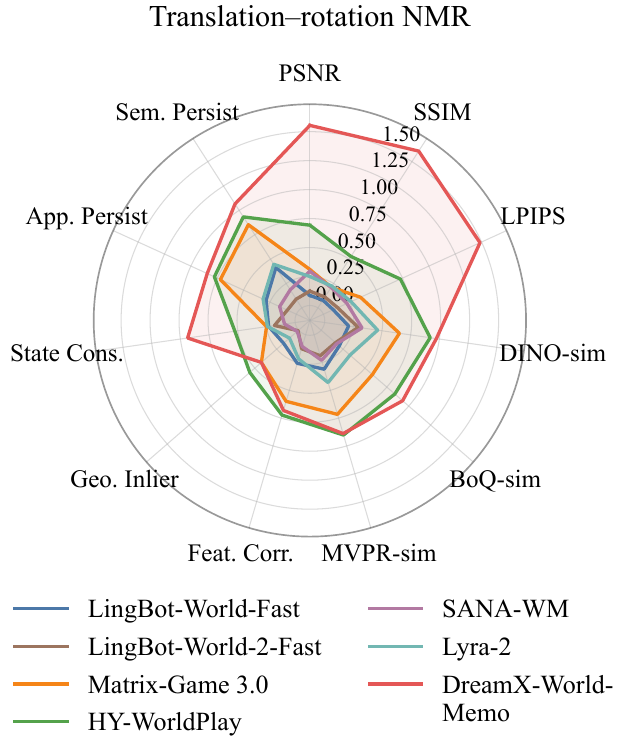}\\[-0.5em]
{\small (b) Translation--rotation}
\end{minipage}
\hfill
\begin{minipage}{0.30\textwidth}
\centering
\includegraphics[width=\linewidth]{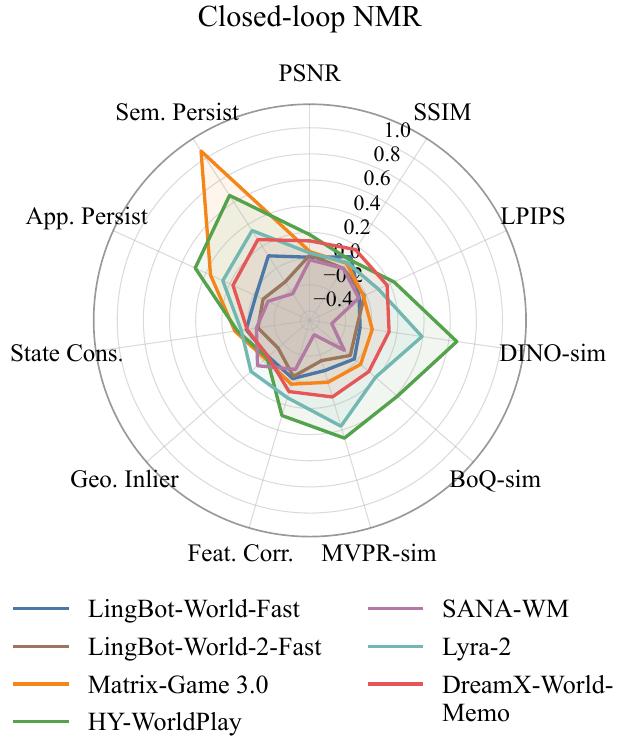}\\[-0.5em]
{\small (c) Closed-loop}
\end{minipage}
\caption{
\textbf{Trajectory-wise metric-level NMR profiles for the video world models.}
Larger values indicate a stronger normalized revisit advantage.
Radial limits are adapted per panel for legibility; compare model shapes and rankings within a panel rather than polygon area across panels.
}
\label{fig:trajectory_nmr_radar}
\end{figure*}

\paragraph{\bf{Trajectory-wise analysis.}}
The three templates probe different forms of return. \emph{Out-and-back} approximately reverses recent motion; \emph{Translation--rotation} adds heading and viewpoint change; and \emph{Closed-loop} requires a return after a longer traversal. Figure~\ref{fig:trajectory_nmr_radar} should therefore be read as three within-template profiles rather than as a direct comparison of radar area.

DreamX-World-Memo has the broadest profile on \emph{Out-and-back} and \emph{Translation--rotation}, but contracts sharply on \emph{Closed-loop}. HY-WorldPlay is less dominant on the first two trajectories yet retains a comparatively broader closed-loop profile, especially on scene and object cues. Matrix-Game 3.0 remains strong on object-related axes but is less uniform across appearance and geometry, while Lyra-2 is intermediate across most dimensions. The two LingBot variants and SANA-WM show limited revisit-selective advantage once the return includes substantial viewpoint change or accumulated rollout error.

The cross-template variation is not a uniform loss of image quality. Some models preserve Appearance Fidelity while Scene Identity or Local Geometry deteriorates; others retain object semantics without recovering a compatible layout. The trajectory suite therefore reveals whether a model supports near-view reinstatement, viewpoint-robust place recognition, or longer-horizon loop closure, rather than reducing all three behaviors to one return score.

\paragraph{\bf{Human validation.}}
Table~\ref{tab:human_score_validation} summarizes the human-alignment and motion-dependence analyses. We evaluate whether the automatic scores track perceived revisit consistency and how much of that agreement can be attributed to motion or camera execution. We sample 10 rollouts from every model--trajectory combination, yielding $7\times3\times10=210$ videos. Six annotators independently rate revisit consistency, generated motion, and camera control from 1 (Bad) to 3 (Good), and we average their ratings for each video.

\begin{table*}[!t]
\centering
\begingroup
\caption{
Human validation on 210 sampled rollouts.
\textbf{(a)} Video-level correlations, including the absolute Raw similarity reference.
$\rho$ is Spearman rank correlation; $H_{\mathrm{cons}}$ and $H_{\mathrm{motion}}$ are human consistency and motion ratings.
Partial $\rho$ controls for camera control, motion, and trajectory.
Action-valid requires mean camera-control and motion ratings both $\geq1.5$.
Raw, MG, and NMR use family-balanced Fisher-$z$ aggregation, whereas Overall NMR is the per-video family-first score.
The 95\% intervals use trajectory--scene cluster bootstrap.
\textbf{(b)} Correlation between each family's NMR and human consistency.
}
\label{tab:human_score_validation}
\small
\setlength{\tabcolsep}{6.0pt}
\renewcommand{\arraystretch}{1.10}
\arrayrulecolor{TableRule}
\scalebox{0.96}{%
\begin{tabular}{@{}lcccccc@{}}
\toprule
\reporttablepanelrow
\multicolumn{7}{c}{\reporttablegroup{(a) Video-level alignment and motion dependence}} \\
\cmidrule(lr){1-7}
\reporttablehead{Score}
& \reporttablehead{$\rho(H_{\mathrm{cons}})$}
& \reporttablehead{95\% CI}
& \reporttablehead{Partial $\rho$}
& \reporttablehead{95\% CI}
& \reporttablehead{Action-valid $\rho$}
& \reporttablehead{$\rho(H_{\mathrm{motion}})$} \\
\midrule
Raw         & 0.594 & $[0.54,0.64]$ & 0.221 & $[0.12,0.30]$ & 0.466 & $-0.336$ \\
MG          & 0.398 & $[0.33,0.46]$ & 0.096 & $[0.00,0.19]$ & 0.280 & $-0.053$ \\
NMR         & 0.480 & $[0.41,0.54]$ & 0.149 & $[0.05,0.25]$ & 0.369 & $-0.119$ \\
\reporttablehighlightrow
Overall NMR & 0.547 & $[0.45,0.63]$ & 0.106 & $[0.02,0.26]$ & 0.409 & $-0.151$ \\
\bottomrule
\end{tabular}
}
\vspace{0.45em}

\setlength{\tabcolsep}{10.5pt}
\scalebox{0.96}{%
\begin{tabular}{@{}lccccc@{}}
\toprule
\reporttablepanelrow
\multicolumn{6}{c}{\reporttablegroup{(b) Per-family NMR alignment with human consistency}} \\
\cmidrule(lr){1-6}
\reporttablehead{Family}
& \reporttablehead{Appearance}
& \reporttablehead{Scene}
& \reporttablehead{Object}
& \reporttablehead{Geometry}
& \reporttablehead{State} \\
\midrule
$\rho(H_{\mathrm{cons}})$
& 0.499
& \textbf{0.618}
& 0.437
& 0.435
& 0.386 \\
\bottomrule
\end{tabular}
}
\arrayrulecolor{black}
\endgroup
\end{table*}

Overall NMR correlates with human consistency at $\rho=0.547$ (95\% CI $[0.45,0.63]$). The metric-level relative scores are also positively correlated with consistency: $\rho=0.398$ for MG and $0.480$ for NMR. These results show that revisit-selective consistency remains perceptually meaningful after calibration against same-video controls.
This alignment is not confined to appearance: the per-family NMR correlations in Table~\ref{tab:human_score_validation}(b) are positive for all five families, ranging from $0.386$ for Persistent State to $0.618$ for Scene Identity.

Table~\ref{tab:human_score_validation} also exposes the trade-off behind this calibration. Raw similarity has a higher unadjusted consistency correlation ($0.594$), but it is substantially coupled to lower motion ($\rho=-0.336$). The corresponding motion correlations are smaller for MG ($-0.053$), NMR ($-0.119$), and Overall NMR ($-0.151$). Relative scoring therefore does not optimize agreement with human ratings in isolation; it retains positive agreement while reducing the slow-motion shortcut that affects absolute similarity.

The complementary objective-motion and controlled diagnostics are reported in Tables~\ref{tab:motion_correlation} and~\ref{tab:confound_analysis}.

\FloatBarrier

\begin{table*}[!t]
\centering
\begingroup
\caption{
Family-level correlations with generated motion. Motion is the mean adjacent-frame optical-flow magnitude computed with RAFT. Correlations are first computed within each model; family cells average their constituent metrics, and Overall weights the five families equally.
}
\label{tab:motion_correlation}
\small
\setlength{\tabcolsep}{3.4pt}
\renewcommand{\arraystretch}{1.10}
\arrayrulecolor{TableRule}
\begin{tabular}{@{}lrrrr@{}}
\toprule
\reporttablepanelrow
\multicolumn{5}{c}{\reporttablegroup{Within-model Spearman $\rho$ with motion}} \\
\cmidrule(lr){1-5}
\reporttablehead{Family} & \reporttablehead{Raw rev.} & \reporttablehead{Temp. base} & \reporttablehead{MG} & \reporttablehead{NMR} \\
\midrule
Appearance Fidelity & $-0.299$ & $-0.304$ & 0.008 & 0.089 \\
Scene Identity      & $-0.274$ & $-0.345$ & 0.185 & 0.063 \\
Local Geometry      & $-0.216$ & $-0.337$ & 0.156 & 0.129 \\
Object Identity     & $-0.061$ & $-0.088$ & 0.090 & 0.023 \\
Persistent State    & $-0.186$ & $-0.283$ & 0.116 & 0.056 \\
\midrule
\reporttablehighlightrow
\textbf{Overall}    & $-0.207$ & $-0.271$ & 0.111 & \textbf{0.072} \\
\bottomrule
\end{tabular}
\arrayrulecolor{black}
\endgroup
\end{table*}

% !TEX root = ../iclr2026_conference.tex

\begin{table*}[!t]
\centering
\begingroup
\newcommand{\bestcell}[1]{\textbf{#1}}
\caption{
Diagnostic analysis of confounds in raw revisit similarity using DreamX-World-Memo generations.
Videos are grouped by motion speed, and visual-quality perturbations are applied to the same High-motion split.
Raw Rev. scores vary with speed and image quality, while Gain and NMR measure revisit-selective consistency after calibration against the same rollout.
\textbf{Bold} denotes the best score within each diagnostic block.
}
\label{tab:confound_analysis}
\scriptsize
\setlength{\tabcolsep}{3.2pt}
\renewcommand{\arraystretch}{1.22}
\arrayrulecolor{TableRule}
\resizebox{\textwidth}{!}{%
\begin{tabular}{lrrrrrrrrrrrrrrrr}
\toprule
\reporttablepanelrow
& \multicolumn{4}{c}{\reporttablegroup{SSIM $\uparrow$}}
& \multicolumn{4}{c}{\reporttablegroup{LPIPS $\downarrow$}}
& \multicolumn{4}{c}{\reporttablegroup{DINO-sim $\uparrow$}}
& \multicolumn{4}{c}{\reporttablegroup{Geo. Inlier $\uparrow$}} \\
\cmidrule(lr){2-5}
\cmidrule(lr){6-9}
\cmidrule(lr){10-13}
\cmidrule(lr){14-17}
\reporttablehead{Setting} & \reporttablehead{Rev.} & \reporttablehead{Base.} & \reporttablehead{Gain $\uparrow$} & \reporttablehead{NMR}
& \reporttablehead{Rev.} & \reporttablehead{Base.} & \reporttablehead{Gain $\uparrow$} & \reporttablehead{NMR}
& \reporttablehead{Rev.} & \reporttablehead{Base.} & \reporttablehead{Gain $\uparrow$} & \reporttablehead{NMR}
& \reporttablehead{Rev.} & \reporttablehead{Base.} & \reporttablehead{Gain $\uparrow$} & \reporttablehead{NMR} \\
\midrule
\reporttablegrouprow
\multicolumn{17}{c}{\reporttablegroup{\textit{Motion speed split}}} \\
\midrule
Low-motion
& \bestcell{0.507} & \bestcell{0.445} & +0.062 & 0.39
& \bestcell{0.318} & \bestcell{0.444} & \bestcell{+0.126} & \bestcell{0.47}
& \bestcell{0.916} & \bestcell{0.861} & +0.055 & \bestcell{0.48}
& \bestcell{0.829} & \bestcell{0.815} & +0.014 & 0.12 \\
Mid-motion
& 0.468 & 0.404 & \bestcell{+0.064} & \bestcell{0.43}
& 0.437 & 0.509 & +0.072 & 0.39
& 0.872 & 0.803 & +0.069 & 0.38
& 0.770 & 0.746 & +0.025 & \bestcell{0.17} \\
High-motion
& 0.452 & 0.408 & +0.044 & 0.33
& 0.458 & 0.549 & +0.091 & 0.31
& 0.691 & 0.620 & \bestcell{+0.071} & 0.42
& 0.722 & 0.693 & \bestcell{+0.029} & 0.16 \\
\midrule
\reporttablegrouprow
\multicolumn{17}{c}{\reporttablegroup{\textit{Quality perturbation on the same High-motion split}}} \\
\midrule
High-blur
& \bestcell{0.550} & \bestcell{0.497} & \bestcell{+0.053} & \bestcell{0.37}
& 0.430 & 0.526 & \bestcell{+0.096} & \bestcell{0.33}
& 0.685 & 0.612 & \bestcell{+0.073} & \bestcell{0.22}
& \bestcell{0.705} & \bestcell{0.673} & \bestcell{+0.032} & 0.16 \\
High-downup
& 0.545 & 0.492 & \bestcell{+0.053} & 0.36
& \bestcell{0.429} & \bestcell{0.523} & +0.094 & \bestcell{0.33}
& \bestcell{0.687} & \bestcell{0.614} & \bestcell{+0.073} & \bestcell{0.22}
& 0.698 & 0.665 & \bestcell{+0.032} & \bestcell{0.17} \\
High-comp.
& 0.451 & 0.406 & +0.045 & 0.33
& 0.463 & 0.552 & +0.089 & 0.31
& 0.683 & 0.612 & +0.071 & \bestcell{0.22}
& 0.682 & 0.656 & +0.026 & 0.13 \\
\bottomrule
\end{tabular}
}
\arrayrulecolor{black}
\endgroup
\end{table*}

The association remains positive under two controls. After rank-residualizing camera control, motion, and trajectory, NMR has $\rho=0.149$ (95\% CI $[0.05,0.25]$), and Overall NMR has $\rho=0.106$ (95\% CI $[0.02,0.26]$). On the action-valid subset, which excludes videos with the clearest motion or camera-control failures, the correlations rise to $0.369$ and $0.409$. Together, these analyses support perceptual alignment of the observable score without treating motion, action execution, or an internal memory mechanism as interchangeable with revisit consistency. Appendix~\ref{app:human_validation} reports inter-rater reliability, per-family results, and clustering sensitivity.

\paragraph{\bf{Objective motion diagnostic.}}
Human ratings capture perceived motion, but we also test motion dependence with mean adjacent-frame optical-flow magnitude from RAFT~\citep{teed2020raft}. To avoid conflating motion with model identity, we compute Spearman correlations within each model and aggregate them by metric family. Raw denotes the absolute base-metric value on revisit pairs, and Temp.\ base denotes ordinary same-video similarity at a matched temporal gap. Table~\ref{tab:motion_correlation} reports the family-level correlations.

Raw revisit similarity and the temporal baseline are negatively correlated with motion in every family: videos with less generated motion preserve higher generic similarity. After family-first aggregation, the correlation is $-0.207$ for Raw and $-0.271$ for the baseline, compared with $0.111$ for MG and $0.072$ for NMR. Relative calibration therefore removes most of the systematic slow-motion advantage visible in absolute similarity. The remaining positive correlations in several families also show the boundary of this result: MG and NMR are less motion-sensitive, not motion-invariant, and nearly static or action-invalid outputs should still be inspected separately. Appendix~\ref{app:full_motion} gives all 11 metric-level correlations and confidence indicators.

\paragraph{\bf{Controlled diagnostic of relative scoring.}}
We complement the across-video correlations with two controlled comparisons on DreamX-World-Memo generations. The first groups natural rollouts by realized motion speed; the second applies blur, downsample--upsample, and compression to the same high-motion subset. Table~\ref{tab:confound_analysis} reports both comparisons.

\FloatBarrier

Across the natural motion split in Table~\ref{tab:confound_analysis}, raw SSIM decreases from 0.507 to 0.452, DINO similarity from 0.916 to 0.691, and geometric inlier ratio from 0.829 to 0.722 as motion increases. MG remains positive for every displayed metric and speed group, but is not required to vary monotonically because its matched baseline changes with the rollout. The relative scores therefore test revisit-selective recovery rather than treating the slowest generation as the strongest one.

The quality perturbations provide a same-video comparison: blur, downsample--upsample, and compression are applied to the same high-motion subset. Raw SSIM ranges from 0.451 to 0.550, while SSIM MG remains within 0.045--0.053; LPIPS MG remains within 0.089--0.096. Other metrics show smaller raw shifts, and geometry retains some residual variation. Together with the human and optical-flow analyses, these controlled results support attenuation, rather than elimination, of shortcuts caused by low motion and globally shared rendering quality.

% !TEX root = ../iclr2026_conference.tex

\section{Conclusion}
R2M-Bench measures whether an interactive video world model is selectively more consistent at commanded revisits than at gap-matched, non-target moments in the same rollout. MG reports direct metric-level revisit advantage, while NMR normalizes this advantage for family-level and overall aggregation across five dimensions. Video-level human judgments align with the benchmark, and independent optical-flow diagnostics show substantially reduced motion dependence after relative calibration. Results reveal dimension- and trajectory-specific gaps, with retrieval-based systems leading among the evaluated video models.

\section{Limitations}
R2M-Bench evaluates observable revisit-selective consistency rather than identifying an internal memory mechanism; retrieval-related findings are associative, not causal. Revisit mining relies on commanded poses and model-specific action interfaces, so execution errors can remain entangled with generated-world persistence even though the human and motion analyses control for important manifestations of this confound. Automatic metrics also inherit backbone, viewpoint, and prompt biases. Finally, the current navigation-only scope excludes object interaction and deliberately evolving state, motivating realized-trajectory evaluation and broader interactive tasks.

\ifarxivversion\else
\input{sections/acknowledgments}
\fi

\begingroup
\setlength{\bibsep}{4pt}
\bibliography{iclr2026}
\bibliographystyle{plainnat}
\endgroup

\clearpage
\appendix
% !TEX root = ../iclr2026_conference.tex

\flushbottom

\section{Implementation and Evaluation Details}

This appendix first specifies the metric implementations and vision-language rubric, then explains the dynamic-range normalization used by NMR. It next provides the complete human and motion analyses underlying the compact validation tables in the main text, followed by controlled diagnostics and family-wise rank comparisons.

\subsection{Metric Implementation Details}
\label{app:metric_details}

This section gives implementation details for the metric families in Section~\ref{sec:method}. All metrics are computed on individual frame pairs and then averaged over the revisit, baseline, and short-range pair sets before computing MemoryGain and NMR. Unless otherwise specified, RGB frames are evaluated at their decoded video resolution.

\paragraph{Appearance Fidelity.}
For a frame pair $(f_i,f_j)$, PSNR is computed with pixel range 255:
\begin{equation}
\mathrm{PSNR}(f_i,f_j)
=
10\log_{10}\frac{255^2}{\mathrm{MSE}(f_i,f_j)}.
\end{equation}
SSIM is computed on RGB images with data range 255, channel-wise image layout, and a $7\times7$ local window. LPIPS is computed with the AlexNet LPIPS network. Before LPIPS evaluation, each RGB image is converted from uint8 values in $[0,255]$ to a tensor in $[-1,1]$. PSNR and SSIM are higher-is-better metrics, while LPIPS is lower-is-better and is direction-flipped when computing MemoryGain and NMR.

\paragraph{Scene Identity Preservation.}
For DINOv2, each image is resized to 518 pixels, center-cropped to $518\times518$, normalized with ImageNet statistics, and passed through a DINOv2 ViT-B/14 encoder. The resulting global feature is $\ell_2$-normalized. The DINOv2 score is cosine similarity:
\begin{equation}
q_{\mathrm{DINO}}(f_i,f_j)
=
\frac{\phi_{\mathrm{DINO}}(f_i)^\top\phi_{\mathrm{DINO}}(f_j)}
{\|\phi_{\mathrm{DINO}}(f_i)\|_2\|\phi_{\mathrm{DINO}}(f_j)\|_2}.
\end{equation}
BoQ is evaluated with its DINOv2-backed VPR model. Images are resized to $322\times322$, normalized with ImageNet statistics, encoded into BoQ descriptors, $\ell_2$-normalized, and compared by cosine similarity. MutualVPR follows the same structure: images are resized to $518\times518$, normalized with ImageNet statistics, encoded by the MutualVPR model into a 512-dimensional descriptor, $\ell_2$-normalized, and compared by cosine similarity. All three Scene Identity Preservation scores are higher-is-better.

\paragraph{Object Identity.}
The object-centric metric is computed by pairwise re-detection rather than long-range tracking.
For the first frame $f_i$, we use GroundingDINO with a generic object prompt covering common scene entities, including furniture, buildings, vehicles, plants, signs, doors, windows, and other object categories.
Detections are filtered with box threshold 0.25, text threshold 0.20, non-maximum suppression IoU 0.60, and a minimum SAM2 mask area ratio of 0.002.
SAM2 is then used to refine each retained box into a mask.
We rank detections by GroundingDINO confidence times mask area ratio and keep the top $K=10$ first-frame anchors.

Let $\mathcal{A}_i$ be the retained anchor objects in $f_i$.
For each anchor $o\in\mathcal{A}_i$, we normalize its detected phrase into an object concept $c_o$ and re-detect that concept in the paired frame $f_j$ using the prompt ``$c_o$ .''.
We keep at most five paired-frame candidates per concept and refine their boxes with SAM2.
For each anchor and candidate, pixels outside the SAM2 mask are replaced by white before extracting crop features, so the object score is less affected by background changes.
The appearance score between anchor $o$ and candidate $k$ is the rescaled cosine similarity of their masked DINOv2 crop features:
\begin{equation}
a_{ok}
=
\operatorname{clip}_{01}
\left(
\frac{
\phi_{\mathrm{DINO}}(o)^\top \phi_{\mathrm{DINO}}(k)
}{
2\|\phi_{\mathrm{DINO}}(o)\|_2\|\phi_{\mathrm{DINO}}(k)\|_2
}
+\frac{1}{2}
\right),
\end{equation}
where $\operatorname{clip}_{01}(x)=\min(1,\max(0,x))$.
The semantic score combines the GroundingDINO confidence $g_k$ for the candidate concept with a CLIP~\citep{radford2021clip} image-text score.
Specifically, we compare the candidate crop with the anchor concept text and average the two semantic signals:
\begin{equation}
u_{ok}
=
\frac{1}{2}\operatorname{clip}_{01}(g_k)
+\frac{1}{2}\operatorname{clip}_{01}\left(
\frac{
\psi_{\mathrm{img}}(k)^\top\psi_{\mathrm{text}}(c_o)
}{
2\|\psi_{\mathrm{img}}(k)\|_2\|\psi_{\mathrm{text}}(c_o)\|_2
}
+\frac{1}{2}
\right).
\end{equation}
The implementation falls back to $u_{ok}=\operatorname{clip}_{01}(g_k)$ if no CLIP checkpoint is provided, but all reported object-centric results use CLIP.
The candidate match score is
\begin{equation}
r_{ok}
=
\frac{
w_{\mathrm{app}}a_{ok}+w_{\mathrm{sem}}u_{ok}
}{
w_{\mathrm{app}}+w_{\mathrm{sem}}
},
\end{equation}
with default weights $w_{\mathrm{app}}=w_{\mathrm{sem}}=0.5$.
The best candidate is accepted as a match if $\max_k r_{ok}\geq0.50$; otherwise the anchor is considered unmatched.

Let $\mathcal{M}_{ij}\subseteq\mathcal{A}_i$ be the matched anchors, and let $k(o)$ denote the accepted candidate for anchor $o$.
If no first-frame anchor is retained, all object-centric scores for the pair are set to 0.
We compute two matched-only identity scores,
\begin{equation}
q_{\mathrm{obj\text{-}app}}(f_i,f_j)
=
\begin{cases}
\frac{1}{|\mathcal{M}_{ij}|}\sum_{o\in\mathcal{M}_{ij}} a_{o,k(o)}, & |\mathcal{M}_{ij}|>0,\\
0, & |\mathcal{M}_{ij}|=0,
\end{cases}
\end{equation}
\begin{equation}
q_{\mathrm{obj\text{-}sem}}(f_i,f_j)
=
\begin{cases}
\frac{1}{|\mathcal{M}_{ij}|}\sum_{o\in\mathcal{M}_{ij}} u_{o,k(o)}, & |\mathcal{M}_{ij}|>0,\\
0, & |\mathcal{M}_{ij}|=0.
\end{cases}
\end{equation}
We also compute persistence scores over all first-frame anchors, assigning zero to unmatched anchors:
\begin{equation}
q_{\mathrm{app\text{-}persist}}(f_i,f_j)
=
\frac{1}{|\mathcal{A}_i|}
\sum_{o\in\mathcal{A}_i}
\mathbf{1}[o\in\mathcal{M}_{ij}]\,a_{o,k(o)},
\end{equation}
\begin{equation}
q_{\mathrm{sem\text{-}persist}}(f_i,f_j)
=
\frac{1}{|\mathcal{A}_i|}
\sum_{o\in\mathcal{A}_i}
\mathbf{1}[o\in\mathcal{M}_{ij}]\,u_{o,k(o)}.
\end{equation}
The main tables report the two persistence metrics, denoted App. Persist and Sem. Persist, because they penalize both object disappearance and identity degradation.
All object metrics are higher-is-better and enter the same MemoryGain and NMR computation as the other pairwise metrics.

\paragraph{Local Geometric Correspondence.}
We detect local features using SuperPoint with at most 1024 keypoints per image and detection threshold $5\times10^{-4}$. Keypoints are matched using LightGlue with SuperPoint features and matching filter threshold 0.1. Let $K_i$ and $K_j$ be the numbers of detected keypoints in the two images and let $M_{ij}$ be the set of LightGlue matches. We report the feature-correspondence ratio
\begin{equation}
q_{\mathrm{match}}(f_i,f_j)
=
\frac{|M_{ij}|}{\max(\min(K_i,K_j),1)}.
\end{equation}
To measure geometric compatibility of the matched points, we estimate a fundamental matrix with OpenCV RANSAC. Each matched keypoint is represented in homogeneous image coordinates as $\mathbf{x}_i=(u_i,v_i,1)^\top$ and $\mathbf{x}_j=(u_j,v_j,1)^\top$. A valid two-view fundamental matrix $F$ should satisfy the epipolar constraint
\begin{equation}
\mathbf{x}_j^\top F\mathbf{x}_i = 0,
\end{equation}
where $F\mathbf{x}_i$ is the epipolar line in the second image and $F^\top\mathbf{x}_j$ is the corresponding epipolar line in the first image. In RANSAC, OpenCV repeatedly samples candidate correspondence subsets, estimates a rank-2 fundamental matrix from the linear epipolar equations, and counts as inliers the matches whose point-to-epipolar-line error is below 3.0 pixels. The confidence parameter is set to 0.99. If fewer than eight matches are available, or if robust estimation fails, the RANSAC inlier ratio is set to 0. Otherwise, with inlier set $I_{ij}\subseteq M_{ij}$, we report
\begin{equation}
q_{\mathrm{inlier}}(f_i,f_j)
=
\frac{|I_{ij}|}{|M_{ij}|}.
\end{equation}
Both local-geometry metrics are higher-is-better. Although the evaluation code can also compute Sampson residual diagnostics, those residual metrics are not used in the reported tables.

\subsection{Vision-Language Evaluation Prompt}
\label{app:vlm_prompt}

We apply the same fixed prompt to every frame pair in the revisit, baseline, and short-range sets. The vision-language model is not told which set a pair comes from; it assigns a pairwise persistent-state consistency score using only the two images. The rubric first asks whether the images contain evidence of the same local scene, then accounts for viewpoint and occlusion when judging shared content. This distinction is important because a pair with no recognizable shared scene should receive a low consistency score rather than a perfect score merely because no local contradiction can be identified. The full prompt is shown below.

\begin{Verbatim}[breaklines=true,breakanywhere=false,breaksymbolleft={},breaksymbolright={},fontsize=\scriptsize]
You are an expert evaluator of Video World Models.

Task: Assign a pairwise persistent-state consistency score to two images
sampled from the same generated video.

* Image A is earlier and Image B is later.
* Do NOT assume that the images depict the same place.
* This is not raw pixel similarity: viewpoint, occlusion, and mild rendering
  changes may differ without implying a state inconsistency.
* Score whether the images provide evidence of the same local scene and, when
  they do, whether the shared scene content remains persistent.

Procedure:

1. Classify the scene relation as same, partial, different, or uncertain.
2. If shared scene content exists, estimate the viewpoint change and identify
   the co-visible region.
3. Judge identity, geometry, and persistence only where content can be
   compared, discounting differences explained by viewpoint or occlusion.
4. If there is no recognizable shared scene content, assign low consistency;
   absence of comparable content is not evidence of perfect consistency.

Viewpoint exemptions for same or partially overlapping scenes: do NOT penalize:
* Any difference plausibly caused by camera motion, perspective, scale, or
  occlusion.
* Objects entering or leaving the frame due to the viewpoint shift.
* Minor lighting, exposure, or rendering-noise differences.
* When uncertain whether a difference is from viewpoint or a true change,
  assume viewpoint.

Inconsistency types (report only confident findings):
* scene: images depict different local scenes or share no recognizable place cues.
* identity: same object changed color, shape, material, texture, or category.
* geometry: impossible spatial layout or structural contradictions.
* persistence: object disappeared or appeared well inside the shared region,
  or count of repeated elements changed without occlusion explanation.

Scoring anchors:
* 5.0 = strong evidence of the same local scene; shared state is consistent.
* 4.0 = same local scene with only minor confident inconsistencies.
* 3.0 = partial or uncertain scene overlap, or moderate inconsistencies.
* 2.0 = weak shared-scene evidence or major state inconsistencies.
* 1.0 = clearly different local scenes or fundamentally incompatible states.
Intermediate values are allowed. Do not give a high score solely because
the images are individually plausible or because no content is comparable.

Output constraints: keep SHORT:
* scene_relation is one of: same, partial, different, uncertain
* viewpoint_difference ≤ 15 words
* at most 3 inconsistencies, each description ≤ 12 words
* reason ≤ 15 words
* Each inconsistency has ONLY keys: type, severity, description.

Output ONLY this JSON (no markdown):
{
  "scene_relation": "same|partial|different|uncertain",
  "viewpoint_difference": "...",
  "inconsistencies": [
    {
      "type": "scene|identity|geometry|persistence",
      "severity": "minor|moderate|major|catastrophic",
      "description": "..."
    }
  ],
  "score": 0.0,
  "reason": "..."
}
\end{Verbatim}

\begin{figure*}[!t]
\centering
\makebox[\textwidth][c]{%
\includegraphics[
    width=1.08\textwidth,
    trim=24 14 24 16,
    clip
]{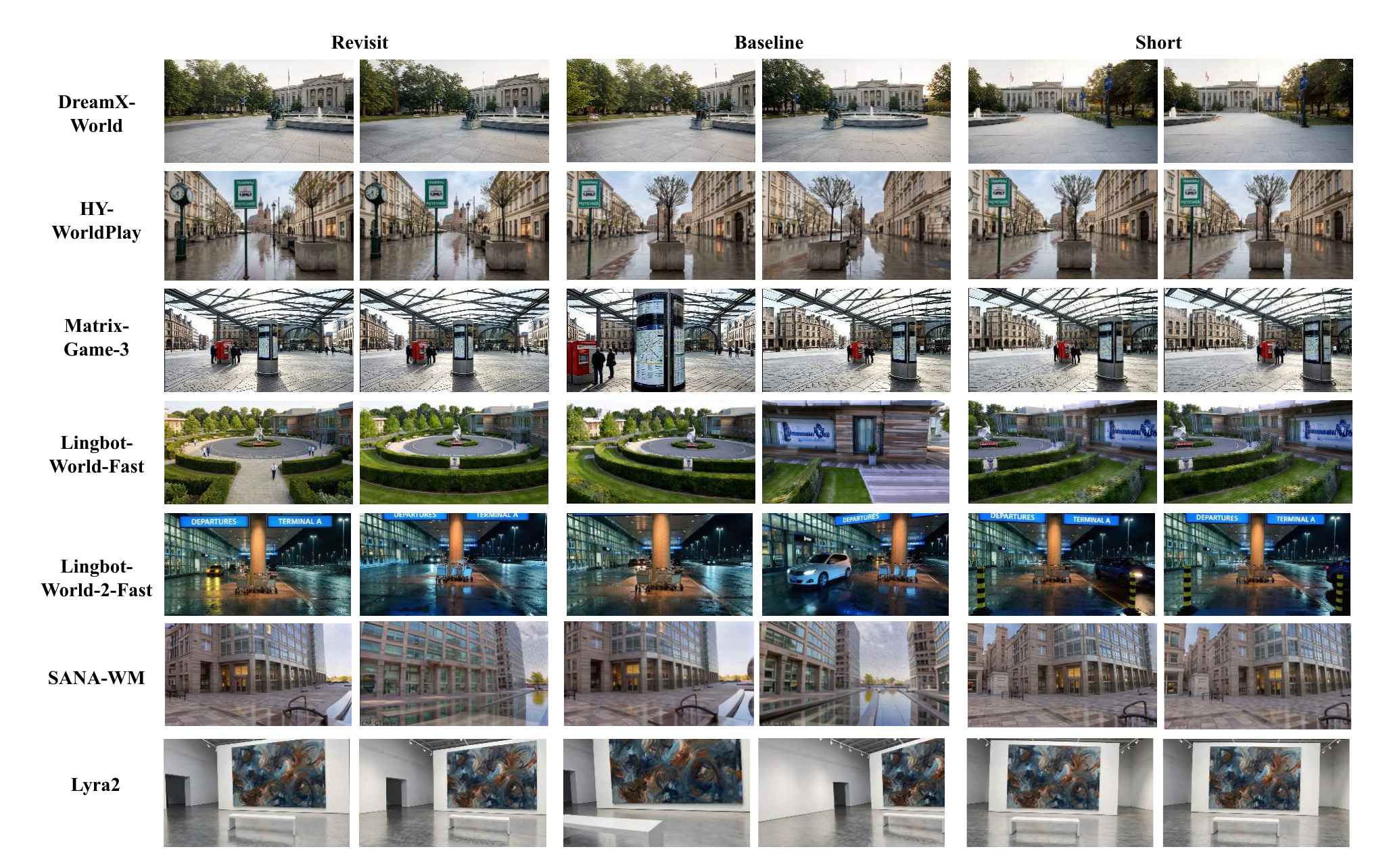}%
}
\caption{
\textbf{Qualitative examples of the three pair types used by the relative protocol.}
Each row shows one model; columns show a revisit pair, a gap-matched non-revisit baseline, and a short-range reference from the same rollout.
The protocol asks whether the revisit recovers more shared scene evidence than the long-gap baseline, calibrated by the consistency observed at short range.
}
\label{fig:qualitative_pair_samples}
\end{figure*}

Figure~\ref{fig:qualitative_pair_samples} previews the revisit, gap-matched baseline, and short-range pairs used in the interpretation below.

% \section{Two-Template NMR Summary}
% \label{app:two_template_nmr}

% Table~\ref{tab:main_nmr_two_template_summary} reports a supplemental NMR summary averaged over only the \emph{Out-and-back} and \emph{Translation--rotation} trajectory templates.
% This analysis excludes the Closed-loop template and keeps the same displayed metric set and Overall averaging rule as Table~\ref{tab:main_nmr_summary}.

% \input{tables/two_template_nmr_summary}

\section{Interpreting NMR and Its Dynamic-Range Dependence}
\label{app:score_interpretation}

NMR is the dimensionless score used for cross-metric aggregation, while MemoryGain (MG) is the direct metric-level revisit advantage. NMR divides MG by the short-range dynamic range (DR), which measures the consistency gap between short-range and baseline pairs. This normalization enables comparison across numerical scales but depends on a reliable denominator: when DR is small, NMR can be unstable or inflated. Overall NMR supports aggregate comparison, but individual values and close rankings should therefore be interpreted with MG and DR.
Table~\ref{tab:mg_dr_nmr_interpretation} summarizes typical diagnostic cases.

\par\medskip
\noindent\begin{minipage}{\textwidth}
\centering
\captionsetup{hypcap=false}
\captionof{table}{
Diagnostic interpretation of NMR through its numerator MG and denominator DR.
``High'' and ``low'' are interpreted relative to the empirical scale of the corresponding metric and trajectory.
}
\label{tab:mg_dr_nmr_interpretation}
\scriptsize
\setlength{\tabcolsep}{4pt}
\renewcommand{\arraystretch}{1.10}
\arrayrulecolor{TableRule}
\begin{tabular}{@{}p{0.12\linewidth}p{0.12\linewidth}p{0.16\linewidth}p{0.52\linewidth}@{}}
\toprule
\reporttablehead{MG} & \reporttablehead{DR} & \reporttablehead{NMR} & \reporttablehead{Diagnostic meaning} \\
\midrule
High & High & High
& Reliable high retention. The revisit is clearly better than baseline and recovers a large fraction of the model's short-range consistency. \\

High & High & Low / Moderate
& Partial retention. The revisit has a clear advantage over baseline, but still falls short of the model's short-range consistency. \\

High & Low & High
& Caution case. The revisit advantage may be real, but NMR can be amplified because the short-range normalization scale is weak. \\

Low & High & Low
& Revisit failure under a reliable scale. The metric can separate short-range pairs from baseline pairs, but the revisit gains little over baseline. \\

Low & Low & Unstable
& Weak diagnostic regime. The metric provides little short-range separation, so the normalized NMR value is not reliable. \\

Negative & Positive & Negative
& Negative revisit effect. The revisit is less consistent than gap-matched non-revisit controls, indicating severe drift or scene/state replacement. \\

Positive & Near-zero / Negative & Unstable
& Invalid or unreliable normalization. Short-range consistency is not clearly above baseline, so NMR is not reliable for aggregation. \\
\bottomrule
\end{tabular}
\arrayrulecolor{black}
\end{minipage}
\par\medskip

At the boundaries, $\mathrm{NMR}\leq0$ means that the revisit does not outperform the baseline, with negative values indicating worse consistency than long-gap controls. $\mathrm{NMR}\approx1$ reaches the metric's short-range level, while a value above one exceeds it. Values above one are convincing only with positive MG and meaningful DR; otherwise they may reflect an unstable denominator. Figure~\ref{fig:qualitative_pair_samples} illustrates the pair types underlying this comparison.

\section{Human Evaluation and Statistical Validation}
\label{app:human_validation}

\subsection{Protocol}

\begin{figure*}[!t]
    \centering
    \includegraphics[width=\textwidth]{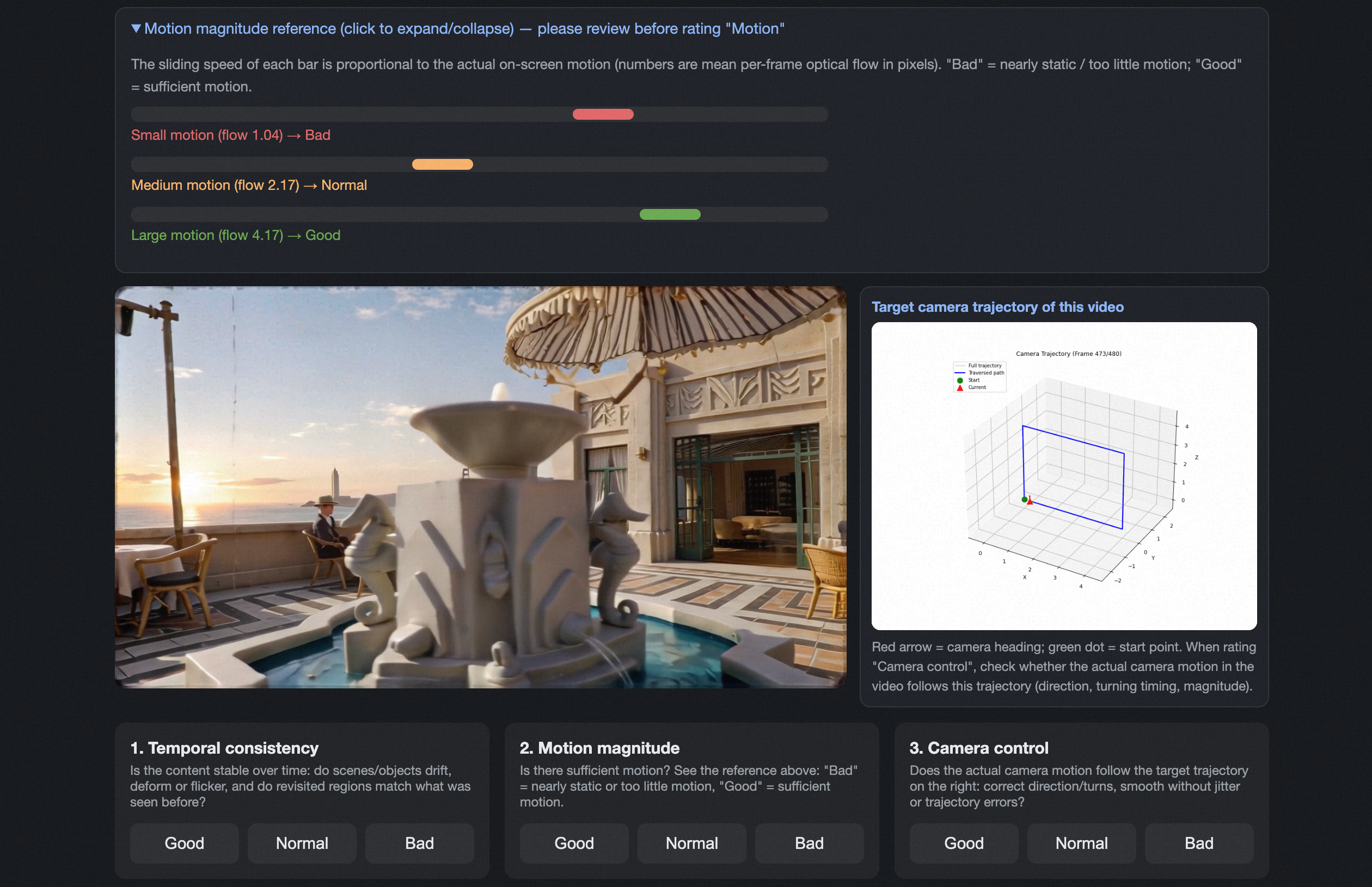}
    \caption{\textbf{Human annotation interface used in our study.} Annotators view the generated video together with the target camera trajectory and motion-magnitude references, then independently rate temporal/revisit consistency, motion magnitude, and camera control as Bad, Normal, or Good. The separate controls make it possible to distinguish perceived revisit inconsistency from insufficient motion or failure to follow the commanded trajectory.}
    \label{fig:human_annotation_interface}
\end{figure*}

The human study samples 10 rollouts for each of the seven video world models and three trajectory templates, yielding $7\times3\times10=210$ videos. Six annotators independently score each video along three axes: revisit consistency, generated motion, and camera control. Each axis uses an ordinal scale with 1 (Bad), 2 (Normal), and 3 (Good), and we average the six ratings per video. Human consistency is the target judgment; motion and camera control record whether a model visibly moves and follows the commanded path rather than remaining static or drifting arbitrarily. Figure~\ref{fig:human_annotation_interface} documents the interface presented to the annotators. Alongside the generated video, it provides the commanded camera trajectory and optical-flow-calibrated motion references, making the nuisance judgments concrete rather than asking annotators to infer them from verbal instructions alone.

We define an \emph{action-valid} subset to test the benchmark after excluding the clearest control failures. A video is action-valid when both its mean camera-control rating and mean motion rating are at least 1.5. Inter-rater reliability, measured with ordinal Krippendorff's $\alpha$, is 0.339 for consistency, 0.371 for motion, and 0.527 for camera control. The modest agreement on revisit consistency reflects the perceptual difficulty of deciding whether a long generated rollout recovers the same scene rather than a merely similar one; correlation magnitudes should therefore be interpreted with this annotation uncertainty in mind.

\FloatBarrier

\subsection{Score Construction and Uncertainty}

For Raw, MG, and NMR, we first compute the Spearman correlation between each benchmark metric and the corresponding human rating across the 210 videos. Because correlations are dimensionless but not directly additive, we transform each metric-level correlation with Fisher's $z=\operatorname{arctanh}(\rho)$, average within each of the five metric families, weight the five family means equally, and apply $\tanh$ to return to the correlation scale. Thus, a family contributes one fifth of the aggregate regardless of how many metrics it contains. Overall NMR follows a different but consistent route: we construct the family-first Overall NMR for each video and correlate that single per-video score directly with the human ratings.

Partial correlations use rank residualization. We rank the score and human-consistency variables, regress both on human-rated camera control, human-rated motion, and trajectory identity, and correlate the residuals. This tests whether the score retains an association with perceived consistency beyond these observed nuisance variables; it does not remove unmeasured differences in realized viewpoint or action execution.

We compute 95\% confidence intervals with a percentile cluster bootstrap using $B=1000$ replicates. Clusters are keyed by trajectory and reference scene, so outputs from different models sharing the same reference condition remain together. Each replicate resamples these clusters with replacement and recomputes the complete statistic, including the metric-level correlations, Fisher-$z$ family aggregation, and rank residualization for partial correlations. The 2.5th and 97.5th percentiles form the primary interval. As a conservative sensitivity analysis, we also bootstrap the seven model identities as clusters.

\begin{table*}[!t]
\centering
\begingroup
\caption{
Expanded human-validation diagnostics.
\textbf{(a)} Family-level Spearman correlations; each entry aggregates only the benchmark metrics in that family.
\textbf{(b)} Percentile bootstrap intervals under the primary trajectory--scene clustering and a conservative seven-method clustering.
}
\label{tab:app_human_family_sensitivity}
\small
\setlength{\tabcolsep}{4.8pt}
\renewcommand{\arraystretch}{1.10}
\arrayrulecolor{TableRule}
\begin{tabular}{@{}lrrr@{}}
\toprule
\reporttablepanelrow
\multicolumn{4}{c}{\reporttablegroup{(a) Per-family $\rho(H_{\mathrm{cons}})$}} \\
\cmidrule(lr){1-4}
\reporttablehead{Family} & \reporttablehead{Raw} & \reporttablehead{MG} & \reporttablehead{NMR} \\
\midrule
Appearance Fidelity & 0.454 & 0.487 & 0.499 \\
Scene Identity      & 0.733 & 0.457 & 0.618 \\
Local Geometry      & 0.651 & 0.364 & 0.435 \\
Persistent State    & 0.598 & 0.332 & 0.386 \\
Object Identity     & 0.482 & 0.339 & 0.437 \\
\bottomrule
\end{tabular}
\hspace{0.06\textwidth}
\begin{tabular}{@{}lcc@{}}
\toprule
\reporttablepanelrow
\multicolumn{3}{c}{\reporttablegroup{(b) 95\% CI sensitivity to clustering}} \\
\cmidrule(lr){1-3}
\reporttablehead{Score} & \reporttablehead{Trajectory--scene} & \reporttablehead{Method} \\
\midrule
Raw         & $[0.54,0.64]$ & $[0.39,0.65]$ \\
MG          & $[0.33,0.46]$ & $[0.19,0.48]$ \\
NMR         & $[0.41,0.54]$ & $[0.26,0.56]$ \\
\reporttablehighlightrow
Overall NMR & $[0.45,0.63]$ & $[0.29,0.63]$ \\
\bottomrule
\end{tabular}
\arrayrulecolor{black}
\endgroup
\end{table*}

\subsection{Results and Interpretation}

Table~\ref{tab:human_score_validation} reports the complete Raw, MG, NMR, and Overall NMR comparison. All four scores correlate positively with human consistency under the primary bootstrap. Raw has the largest unadjusted correlation ($0.594$), but also a substantially stronger negative correlation with human-rated motion ($-0.336$) than MG ($-0.053$), NMR ($-0.119$), or Overall NMR ($-0.151$). Absolute similarity therefore aligns with human consistency partly because slowly moving videos are easier to match.

After controlling for camera control, motion, and trajectory, both NMR ($0.149$, 95\% CI $[0.05,0.25]$) and Overall NMR ($0.106$, 95\% CI $[0.02,0.26]$) remain positively correlated with consistency. They also remain positive on the action-valid subset ($0.369$ and $0.409$). These results support perceptual alignment and reduced motion dependence; they do not show that a relative score must have a larger unadjusted human correlation than Raw.

Table~\ref{tab:app_human_family_sensitivity}(a) shows positive human-consistency correlations for every metric family and score type. NMR is strongest for Scene Identity ($0.618$) and Appearance Fidelity ($0.499$), while its correlations for Local Geometry ($0.435$), Object Identity ($0.437$), and Persistent State ($0.386$) show that the alignment is not confined to low-level appearance. Panel (b) repeats the uncertainty analysis with model identity as the clustering unit. The seven-cluster intervals are wider, as expected, but every lower bound remains above zero. We regard this result as a conservative sensitivity check rather than the primary uncertainty estimate.

For completeness, Table~\ref{tab:family_mean_rank} provides the corresponding family-wise rank comparison across score types.

\FloatBarrier

\section{Motion and Controlled-Confound Diagnostics}
\label{app:motion_diagnostics}

\subsection{Full Motion-Correlation Results}
\label{app:full_motion}

\begin{table*}[!t]
\centering
\begingroup
\caption{
Full metric-level motion correlations. Motion is the mean adjacent-frame optical-flow magnitude computed with RAFT. Correlations are computed within each model and aggregated across models with equal weighting using Fisher's $z$ transformation. LPIPS raw and baseline signs are oriented so that higher always indicates better similarity; $^{*}$ denotes a 95\% confidence interval excluding zero.
}
\label{tab:app_motion_correlation_full}
\small
\setlength{\tabcolsep}{7.0pt}
\renewcommand{\arraystretch}{1.06}
\arrayrulecolor{TableRule}
\begin{tabular}{@{}lrrrr@{}}
\toprule
\reporttablepanelrow
\multicolumn{5}{c}{\reporttablegroup{Within-model Spearman $\rho$ with objective motion}} \\
\cmidrule(lr){1-5}
\reporttablehead{Metric} & \reporttablehead{Raw revisit} & \reporttablehead{Temporal baseline} & \reporttablehead{MG} & \reporttablehead{NMR} \\
\midrule
\reporttablegrouprow
\multicolumn{5}{c}{\reporttablegroup{\textit{Appearance Fidelity}}} \\
PSNR  & $-0.242^{*}$ & $-0.235^{*}$ & $-0.063$ & 0.034 \\
SSIM  & $-0.282^{*}$ & $-0.293^{*}$ & 0.045 & 0.109 \\
LPIPS & $-0.373^{*}$ & $-0.384^{*}$ & 0.041 & 0.124 \\
\reporttablegrouprow
\multicolumn{5}{c}{\reporttablegroup{\textit{Scene Identity}}} \\
DINO      & $-0.211^{*}$ & $-0.257^{*}$ & $0.159^{*}$ & 0.049 \\
BoQ       & $-0.342^{*}$ & $-0.417^{*}$ & $0.167^{*}$ & 0.062 \\
MutualVPR & $-0.268^{*}$ & $-0.362^{*}$ & $0.230^{*}$ & $0.078^{*}$ \\
\reporttablegrouprow
\multicolumn{5}{c}{\reporttablegroup{\textit{Local Geometric Correspondence}}} \\
Match ratio   & $-0.259^{*}$ & $-0.372^{*}$ & 0.118 & 0.099 \\
RANSAC inlier & $-0.172^{*}$ & $-0.301^{*}$ & $0.194^{*}$ & $0.159^{*}$ \\
\reporttablegrouprow
\multicolumn{5}{c}{\reporttablegroup{\textit{Object Identity}}} \\
Appearance persistence & $-0.036$ & $-0.094$ & $0.128^{*}$ & 0.084 \\
Semantic persistence   & $-0.085$ & $-0.081^{*}$ & 0.052 & $-0.039$ \\
\reporttablegrouprow
\multicolumn{5}{c}{\reporttablegroup{\textit{Persistent State}}} \\
Gemini score & $-0.186^{*}$ & $-0.283^{*}$ & $0.116^{*}$ & 0.056 \\
\midrule
\reporttablehighlightrow
Median $|\rho|$ & 0.242 & 0.293 & 0.118 & \textbf{0.078} \\
\reporttablehighlightrow
Significant metrics & 9/11 & 10/11 & 6/11 & \textbf{2/11} \\
\bottomrule
\end{tabular}
\arrayrulecolor{black}
\endgroup
\end{table*}

Table~\ref{tab:app_motion_correlation_full} expands the family means in Table~\ref{tab:motion_correlation} into all 11 constituent metrics. Objective motion is the mean adjacent-frame optical-flow magnitude computed with RAFT. To isolate within-model variation, we compute the motion correlation separately for each model and aggregate the correlations across models with equal weighting in Fisher-$z$ space.

Absolute revisit similarity and the temporal baseline retain substantial motion dependence. Their median $|\rho|$ values are 0.242 and 0.293, and their 95\% confidence intervals exclude zero for 9/11 and 10/11 metrics, respectively. The pattern is especially consistent for appearance, scene retrieval, and local geometry, where greater motion makes absolute frame matching harder. For example, BoQ changes from $\rho=-0.342$ for Raw and $-0.417$ for the temporal baseline to $0.062$ for NMR; SSIM changes from $-0.282$ and $-0.293$ to $0.109$.

Relative calibration attenuates this dependence: median $|\rho|$ falls to 0.118 for MG and 0.078 for NMR, and only 2/11 NMR correlations have intervals excluding zero. MutualVPR and RANSAC inliers retain significant positive residual correlations, so NMR is not motion-invariant. Rather, subtracting the same-video temporal baseline removes much of the generic advantage enjoyed by slow or conservative videos while preserving revisit-selective differences.

\begin{table*}[!t]
\centering
\begingroup
\newcommand{\bestRank}[1]{\textbf{#1}}
\newcommand{\secondRank}[1]{\underline{#1}}
\newcommand{\rankTriple}[3]{#1\,/\,#2\,/\,#3}
\caption{
Family-wise mean ranks among the seven video world models.
Each cell reports \textbf{Raw\,/\,MG\,/\,NMR}; lower ranks are better.
The Overall cell averages the corresponding ranks across the five families.
RAFT is mean optical-flow magnitude in pixels per frame and is not ranked.
}
\label{tab:family_mean_rank}
\small
\setlength{\tabcolsep}{3.8pt}
\renewcommand{\arraystretch}{1.18}
\arrayrulecolor{TableRule}
\noindent\resizebox{\textwidth}{!}{
\begin{tabular}{@{}lccccccc@{}}
\toprule
\reporttablepanelrow
\multicolumn{8}{c}{\reporttablegroup{Family-wise mean rank (Raw\,/\,MG\,/\,NMR)}} \\
\midrule
\reporttablehead{Method}
& \reporttablehead{Appearance Fidelity} & \reporttablehead{Scene Identity}
& \reporttablehead{Object Identity} & \reporttablehead{Local Geometry}
& \reporttablehead{Persistent State}
& \reporttablehead{Overall}
& \reporttablehead{\shortstack{RAFT\\px/frame}} \\
\midrule
LingBot-World-Fast
& \rankTriple{6.33}{6.00}{6.33}
& \rankTriple{6.67}{5.00}{5.00}
& \rankTriple{5.50}{5.00}{5.00}
& \rankTriple{6.00}{5.50}{5.00}
& \rankTriple{6.00}{5.00}{5.00}
& \rankTriple{6.10}{5.30}{5.27}
& 2.699 \\

LingBot-World-2-Fast
& \rankTriple{6.33}{6.33}{6.33}
& \rankTriple{5.00}{6.00}{6.00}
& \rankTriple{5.50}{6.50}{6.50}
& \rankTriple{5.00}{6.50}{6.50}
& \rankTriple{5.00}{6.00}{6.00}
& \rankTriple{5.37}{6.27}{6.27}
& 2.482 \\

Matrix-Game 3.0
& \rankTriple{4.00}{4.00}{4.00}
& \rankTriple{4.00}{\secondRank{2.33}}{3.33}
& \rankTriple{\bestRank{1.50}}{\bestRank{1.00}}{\secondRank{2.00}}
& \rankTriple{3.50}{\secondRank{2.50}}{3.00}
& \rankTriple{4.00}{3.00}{3.00}
& \rankTriple{3.40}{2.57}{3.07}
& 2.199 \\

HY-WorldPlay
& \rankTriple{\secondRank{2.00}}{\secondRank{2.00}}{\secondRank{2.00}}
& \rankTriple{\bestRank{1.00}}{2.67}{\secondRank{1.67}}
& \rankTriple{\bestRank{1.50}}{\secondRank{2.50}}{2.50}
& \rankTriple{\bestRank{1.00}}{3.00}{\secondRank{2.00}}
& \rankTriple{\bestRank{1.00}}{\secondRank{2.00}}{\secondRank{2.00}}
& \rankTriple{\bestRank{1.30}}{\secondRank{2.43}}{\secondRank{2.03}}
& 1.038 \\

SANA-WM
& \rankTriple{5.00}{5.67}{5.33}
& \rankTriple{6.33}{7.00}{7.00}
& \rankTriple{7.00}{6.50}{6.50}
& \rankTriple{7.00}{6.00}{6.50}
& \rankTriple{7.00}{7.00}{7.00}
& \rankTriple{6.47}{6.43}{6.47}
& 1.309 \\

Lyra-2
& \rankTriple{3.33}{3.00}{3.00}
& \rankTriple{3.00}{3.67}{3.67}
& \rankTriple{4.00}{4.00}{4.00}
& \rankTriple{3.00}{3.50}{4.00}
& \rankTriple{3.00}{4.00}{4.00}
& \rankTriple{3.27}{3.63}{3.73}
& 2.028 \\

\reporttablehighlightrow
DreamX-World-Memo
& \rankTriple{\bestRank{1.00}}{\bestRank{1.00}}{\bestRank{1.00}}
& \rankTriple{\secondRank{2.00}}{\bestRank{1.33}}{\bestRank{1.33}}
& \rankTriple{\secondRank{3.00}}{\secondRank{2.50}}{\bestRank{1.50}}
& \rankTriple{\secondRank{2.50}}{\bestRank{1.00}}{\bestRank{1.00}}
& \rankTriple{\secondRank{2.00}}{\bestRank{1.00}}{\bestRank{1.00}}
& \rankTriple{\secondRank{2.10}}{\bestRank{1.37}}{\bestRank{1.17}}
& 2.174 \\
\bottomrule
\end{tabular}
}
\arrayrulecolor{black}
\endgroup
\end{table*}

\subsection{Controlled Confound Interpretation}

The across-video analysis above uses natural variation within each model. Table~\ref{tab:confound_analysis} in the main text complements it with controlled DreamX-World-Memo comparisons. Across the natural motion split, absolute revisit scores change markedly: from Low- to High-motion, SSIM falls from 0.507 to 0.452, DINO similarity from 0.916 to 0.691, and geometric inlier ratio from 0.829 to 0.722. MG remains positive for every displayed metric and motion group, but its trend need not be monotonic because the temporal baseline changes with motion as well. This is the rollout-specific nuisance variation that relative scoring is designed to calibrate.

The quality perturbations provide a stricter same-subset comparison: blur, downsample--upsample, and compression are applied to the same High-motion videos. Although absolute SSIM ranges from 0.451 to 0.550, SSIM MG remains within 0.045--0.053 and NMR within 0.33--0.37. LPIPS MG remains within 0.089--0.096, while DINO MG and NMR remain within 0.071--0.073 and at 0.22, respectively. Geometry shows somewhat greater residual variation. These controlled results support attenuation of globally shared quality shifts, not complete invariance to degradation or motion.

\FloatBarrier

\section{Family-wise Rank Comparison}
\label{app:family_rank}

Table~\ref{tab:family_mean_rank} compares score types without averaging incompatible metric units.
For each metric, we rank the seven video world models and then average ranks only within its family.
Within each family cell, ranks are reported in Raw/MG/NMR order; NMR and MG use the oriented higher-is-better convention.
Raw-score ranks use higher-is-better for all metrics except LPIPS.
Each cell reports only the resulting family-wise mean rank; the underlying metric values are reported in Table~\ref{tab:main_nmr_summary}.
The Overall cell then averages the Raw, MG, and NMR ranks, respectively, across the five families so that every family contributes equally.
The RAFT column averages 20 adjacent-frame pairs from each of 20 videos per trajectory (60 videos per model) at a common 320-pixel width.
It measures generated motion magnitude and is reported only to diagnose whether score rankings track slower or faster rollouts.

\end{document}